\documentclass[letterpaper, 10 pt, conference]{ieeeconf}  

\IEEEoverridecommandlockouts                              

\usepackage[T1]{fontenc}
\usepackage{mathptmx} 
\usepackage{times} 
\usepackage{amsmath} 
\usepackage{amssymb}  
\usepackage{graphicx} 
\usepackage[ruled,vlined]{algorithm2e}
\usepackage{booktabs}  
\usepackage{float}
\usepackage{cite} 

\usepackage{bigstrut}
\usepackage{multirow}
\usepackage{balance}
\usepackage[hidelinks,bookmarks=false]{hyperref}  
\usepackage{lineno}
\usepackage{comment}
\usepackage{color}
\usepackage{pbox}
\usepackage{mathtools}
\usepackage{subfigure} 

\usepackage{booktabs}

 \usepackage{multirow}

\usepackage{caption}
 \usepackage{cite}
\usepackage{mathptmx}
\usepackage{mathrsfs}

\begin{document}

\title{R-PCC: A Baseline for Range Image-based Point Cloud Compression }

\author{Sukai Wang, Jianhao Jiao, Peide Cai, and Ming Liu
\thanks{Authors are with The Hong Kong University of Science and Technology, Hong Kong SAR, China. (e-mail: \{swangcy, jjiao, peide.cai\}@connect.ust.hk; eelium@ust.hk). \textit{(Corresponding author: Ming Liu.)}}}

\maketitle

\begin{abstract}
In autonomous vehicles or robots, point clouds from LiDAR can provide accurate depth information of objects compared with 2D images, but they also suffer a large volume of data, which is inconvenient for data storage or transmission. 
In this paper, we propose a \textbf{R}ange image-based \textbf{P}oint \textbf{C}loud \textbf{C}ompression method, R-PCC, which can reconstruct the point cloud with uniform or non-uniform accuracy loss. We segment the original large-scale point cloud into small and compact regions for spatial redundancy and salient region classification. Compared with other voxel-based or image-based compression methods, our method can keep and align all points from the original point cloud in the reconstructed point cloud. It can also control the maximum reconstruction error for each point through a quantization module. In the experiments, we prove that our easier FPS-based segmentation method can achieve better performance than instance-based segmentation methods such as DBSCAN.
To verify the advantages of our proposed method, we evaluate the reconstruction quality and fidelity for 3D object detection and SLAM, as the downstream tasks. The experimental results show that our elegant framework can achieve 30$\times$ compression ratio without affecting downstream tasks, and our non-uniform compression framework shows a great improvement on the downstream tasks compared with the state-of-the-art large-scale point cloud compression methods. Our real-time method is efficient and effective enough to act as a baseline for range image-based point cloud compression. The code is available on https://github.com/StevenWang30/R-PCC.git. 
\end{abstract}

\IEEEpeerreviewmaketitle


\section{Introduction}

Point clouds from scanning LiDAR can not only provide high-accuracy depth information of objects in a large range, but are also suitable for diverse environments like various lighting conditions. However, the large-volume data stream obtained by the LiDAR will lead to problems in practical use, such as storage or transmission. Take Velodyne-64E for example. It will collect 30GB+ of point cloud data per hour. Therefore, large-scale point cloud compression has become necessary for autonomous driving systems. 


Point clouds used in 3D scanning or modeling are very compact and dense. In recent decades, many 3D point cloud compression methods have proposed to compress this type of point cloud, by voxelization \cite{wang2021multiscale,survey} or 3D mesh compression \cite{peng2005technologies}. The point cloud compression ratio will increase when the point cloud becomes denser. 
This is because from the feature aspect, denser points provide more accurate normal information and context features. And from the entropy aspect, the entropy of the point cloud is smaller when the probability of the position of each point is higher.
However, compared with the 3D models, point clouds from scanning LiDAR in outdoor scenarios are commonly scattered through regions up to $100$ meters away.
Thus the compact models can not perform well in the large-scale point cloud compression in the autonomous driving systems. 

Tree-based\cite{GoogleDraco, que2021voxelcontext, biswas2020muscle} and range image-based algorithms\cite{sun2019novel, sun2020novel,sun2020novelseq} are two kinds of effective compression types. Among tree-based methods, in Draco\cite{GoogleDraco}, a mesh and KD-tree are fed into the algorithm for compression, while in \cite{huang2020octsqueeze}, Huang \textit{et al}. firstly reorganized the point cloud into an octree, and used a machine learning model to predict the probability of each voxel, which can be fed into the arithmetic coding algorithm for end-to-end compression. 
Meanwhile, range images are the 2D depth images of 3D point clouds from the frontal view, and an unsynchronized and unrectified point cloud is directly backprojected by the range image stream collected by LiDAR. Image-based compression methods, such as JPEG\cite{wallace1992jpeg} and JPEG2000\cite{rabbani2002jpeg2000}, can be applied to range image compression, such as JPEG or JPEG2000. Compared with the octree-based or KD-tree-based methods, range image-based compression framework can guarantee the number of points in the reconstructed point cloud is exactly the same as that in the original point cloud. Because range image has sharp edges and larger range of values compared with ordinary image from camera, Sun \textit{et al}.\cite{sun2020novel} proposed an instance-based segmentation method that uses Euclidean clustering to reduce the spatial redundancy, and achieved 20$\times$ compression ratio. However, the instance-based or semantic-based segmentation has high time and space complexity for real-time implementation. In this paper, we propose a region-based method using farthest point sampling (FPS). In Sec. \ref{sec:ablation}, we compare the compression ratio and reconstruction quality of instance-based and region-based segmentation methods, and the results illustrate that semantic and accurate segmentation cannot improve the overall compression performance, while our uniform compression framework can achieve 30$\times$ compression ratio with 2cm chamfer distance error. 

Another reason to segment a large-scale point cloud into small regions is that 
we can decrease the compressed bitstream size without affecting downstream tasks by retaining high compression accuracy in important regions and reducing the compression accuracy in unimportant areas. Our non-uniform compression framework uses a key point extractor to classify the clusters into four salience levels. The experimental results for 3D object detection and simultaneous localization and mapping (SLAM) on the KITTI dataset show that our method can achieve a higher compression ratio with less decrease in downstream task performance than other baseline methods.

The contributions of this work are listed as follows:
\begin{itemize}
    \item We evaluate the relationship between compression ratios with different residual ranges and distributions, and the results illustrate that our farthest point sampling segmentation and point-plane mixing modeling method are better than the baseline cluster-based compression method in efficiency and effectiveness.
    \item We propose a uniform and non-uniform compression framework for different requirements. Clusters with more key points keep high reconstruction quality as salient regions for downstream tasks.
    \item We compare our compression framework with other state-of-the-art algorithms, and achieve superior performance in both reconstruction quality and downstream task performance. 
    Our real-time framework, R-PCC, is open-source, easily extended to multiple downstream tasks, and can become a novel range image-based point cloud compression baseline.
\end{itemize}

\begin{figure*}[t]
\centering
\includegraphics[width=0.8\paperwidth]{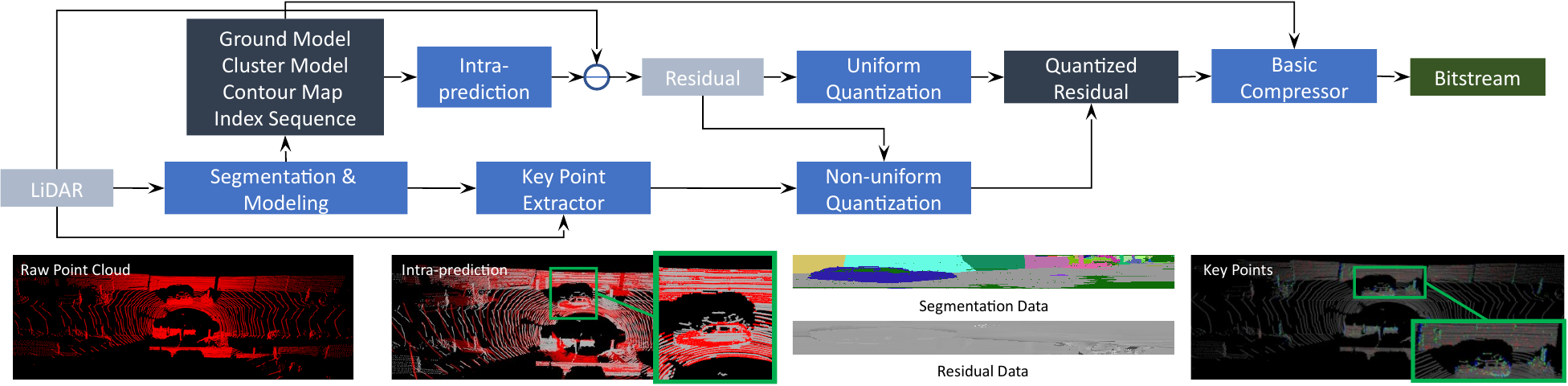}
\caption{Block diagram of our uniform and non-uniform compression framework, R-PCC.  In the non-uniform quantization module, the residual is quantized by different quantization accuracies in each cluster. The segmentation and modeling information data and the quantized residual data are decoded by the basic compressor for binary encoding. }
\label{fig:total}
\vspace{-0.4cm}
\end{figure*}

\section{Related Works}
\subsection{Single-frame Point Cloud Compression}

For static unstructured point clouds, octree representation is commonly utilized as a point-cloud geometry compression method.
Elseberg \textit{et al}. \cite{elseberg2013one} proposed an efficient octree data structure to store and compress 3D data without loss of precision. Experimental results demonstrated that their method was useful for the exchange file format, fast point cloud visualization, fast 3D scan matching, and shape detection.
De Oliveira Renteet \textit{et al}. \cite{de2018graph} introduced a graph-based lossy coding algorithm for the geometry of static point clouds. They used an octree-based technique for the base layer and a graph-based transform technique for the enhancement layer, where a residual was coded, leading to impressive coding performance.
Zhang \textit{et al}. \cite{zhang2017clustering} introduced a clustering and DCT-based color point cloud compression method, in which they used a mean-shift technique to cluster 3D color point clouds into many homogeneous blocks, and a clustering-based prediction method to remove spatial redundancy of the point cloud data. Meanwhile,
Tang \textit{et al}. \cite{tang2016compression} presented an octree-based scattered point cloud compression algorithm, in which the stop condition of segmentation was improved to ensure appropriate voxel size. Additionally, the spatial redundancy and outliers could be removed by traversal queries and bit manipulation.

For static structured point clouds, researchers have focused on employing existing image coders to encode point cloud data by mapping them into 2D arrangements. Houshiar \textit{et al}. \cite{20153D} proposed to project 3D points onto three panorama images and used an image coding method to compress them. Similar to their approach, Ahn \textit{et al}. \cite{ahn2014large} presented an adaptive range image compression algorithm for the geometry information of large-scale 3D point clouds. They explored a prediction method to predict the radial distance of each pixel using previously encoded neighbors and only encoded the resulting prediction residuals. In contrast, Zanuttigh \textit{et al}. \cite{zanuttigh2009compression} focused on efficient compression of RGB-D point cloud data. They developed a segmentation method to identify the edges and main objects of a depth map. After that, an efficient prediction process was performed according to the segmentation result, and the residual data between the predicted and real depth map were calculated. Finally, the few prediction residuals were encoded by conventional image compression methods.
 Sun \textit{et al}.\cite{sun2019novel, sun2020novel,sun2020novelseq}  also proposed a cluster-based method for range images to reduce the residual data range and the entropy of the residual data. However, the ablation and comparative results in this paper shows that the accurate segment methods like DBSCAN for semantic or instance-based segmentation did not show improvement in the overall compression ratio. In our method, we utilize the segmentation results for salience map creation, and we apply a non-uniform framework to achieve a higher compression ratio without a negative effect on the downstream tasks.

\subsection{Key Point Extraction}
For most downstream tasks in autonomous driving systems, the feature extraction of the point cloud is crucial. Feature-based methods focus on the sharp features for edges or flat features for surface matching. LOAM\cite{zhang2014loam} is the standard framework of many current SLAM-related methods like LeGO-LOAM\cite{shan2018lego} or M-LOAM\cite{jiao2021robust}. LOAM splits each scan in the range image into several segments, and finds the top K sharpest and flattest points in the segments as the key points. The more violent the depth change in the adjacent pixels in the same scan, the sharper the edge features. Serafin \textit{et al}.\cite{serafin2016fast} proposed a fast and robust 3D feature extraction method in sparse point clouds, which can extract the planes and  edges.  
In our key point extraction module, we utilize the feature extractor in LOAM\cite{zhang2014loam} and evaluate A-LOAM's \cite{aloam} performance (an open-source implementation of LOAM) the SLAM downstream tasks in the comparative experiments.

\section{Methodology}
\subsection{System Overview}
Our proposed uniform or non-uniform point cloud compression framework R-PCC is shown in Fig. \ref{fig:total}. 
The decompression part of our framework uses the same basic compressor as in the compression framework, to decompress the segmentation and modeling information data (\textit{info. data}) and quantized residual data. The \textit{info. data} can predict the coarse point clouds as in the compression framework, and the residual is recovered by the inverse quantization module. In the non-uniform framework, the accuracy for each cluster corresponds to the quantization module in compression.

Our proposed compression framework is based on the range image. Note that if the range image is collected from the LiDAR, then the number of points is also the same as in the point cloud; and if the range image is projected by the 3D point cloud, the number of points in the reconstructed point cloud depends on the width and height of the range image. 

The accuracy loss thus consists of two parts: 
\begin{enumerate}
    \item The projection from the point cloud to the range image.
    \item The uniform or non-uniform quantization accuracy.
\end{enumerate}

\subsection{Range Image}
\label{sec:tranform}
Currently, single-frame point clouds from most LiDARs can be projected from 3D to 2D. A LiDAR has different laser beams (e.g. Velodyne HDL-64E has 64 lasers, and 32E has 32 lasers), and all the lasers have a full rotation of $360^{\circ}$ along the azimuth direction (horizontal field of view). 

Here we take the Velodyne HDL-64E as an example.
In the altitude direction (vertical field of view),  the range image consists of 64 rows whose angles are distributed between the lowest angle $\varphi_{min}$ and highest angle $\varphi_{max}$. Each scan in the range image represents a fixed angle. If the LiDAR has $H$ lasers, and the horizontal angular resolution is $\rho$, the shape of the range image collected by the LiDAR should be $[H, W] = [H, \lfloor 360 / \rho \rceil]$, where $\lfloor \rceil$ represents the rounding operation. 

We can project a 3D point $\mathbf{P} = (x, y, z)$ onto the corresponding 2D pixel $\mathbf{I} = (w, h, r)$ of a range image, where $w$ and $h$ are the vertical index and horizontal index, and $r$ is the Euclidean distance from the point to the LiDAR origin. 
Values of $(w, h, r)$ are calculated accordingly.
$r = \sqrt{x^2 + y^2 + z^2}$, 
$h = \lfloor \theta / \rho \rceil$, and
$w = \lfloor(\varphi - \varphi_{min}) / \sigma \rceil$,
where $\theta = \arctan(y / x)$ and $\varphi = \arctan(z / r)$ are the 
horizontal angle and vertical angle respectively,
$\varphi_{min}$ is the smallest vertical angle, 
and $\sigma = W / (\varphi_{max} - \varphi_{min})$.



\begin{figure}[t]
\centering
\begin{minipage}[t]{0.48\linewidth}
\centering
\includegraphics[width=1.0\textwidth]{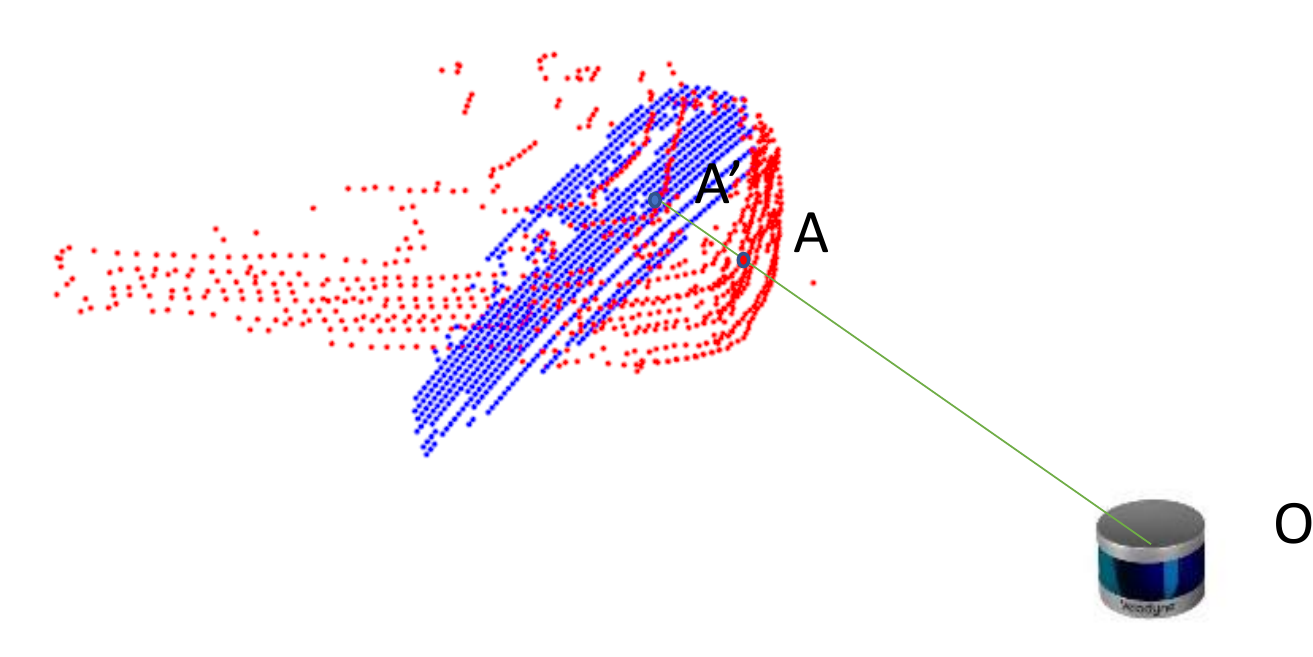}
\end{minipage}
\hfill
\begin{minipage}[t]{0.48\linewidth}
\centering
\includegraphics[width=1.0\textwidth]{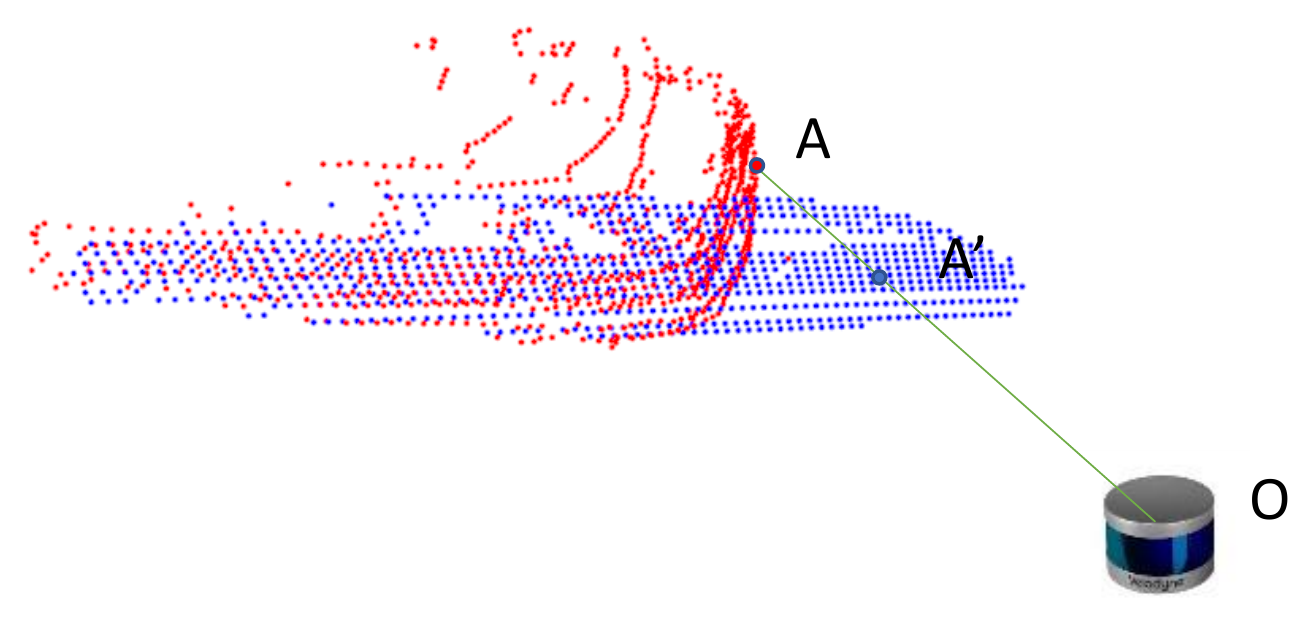}
\end{minipage}
\caption{Example of the point modeling and plane modeling method. The blue points are intra-predicted result using point model or plane model. The residual of point A is OA - OA'.}
\label{fig:modeling_exp}
\vspace{-0.5cm}
\end{figure}

\subsection{Compression Framework}
 
 \textbf{Ground Extraction Module. }The ground points have strong regularity because the ground points can be fitted into a large plane. We estimate the ground model using the RANSAC planer fitting method like \cite{sun2020novel}. 
 
 \textbf{Segmentation Module.} This module segments the point cloud into several denser point cloud subsets. Compared with the instance-based segmentation methods in \cite{sun2019novel} and \cite{sun2020novel}, we choose FPS method to find the center of each cluster as region-based segmentation method. The number of clusters is equal to the number of sampled points setting in FPS. In the Sec. \ref{sec:ablation}, we compared the DBSCAN as baseline with ours segmentation method, and the results shows that ours method performs better in compression and efficiency.
 
 \textbf{Modeling Module.} After obtaining the small point cloud clusters, we use two methods, point and plane, to model the points in each cluster. The point modeling method uses the mean of the points' depth, and the plane modeling method uses a plane, which is estimated by RANSAC, to represent points in each cluster. When the number of points in the cluster is smaller than 30, or the maximum angle between the plane norm and LiDAR scans in this cluster is larger than $75^{\circ}$, we will choose the point modeling method. 
 
 For a point set $\{\mathbf{P}_i=(x_i, y_i, z_i)\}_{i=1}^k$ in the cluster, the point model: $r=\frac{1}{k}\sum ||\mathbf{P}_i||_2$, and the plane model $ax + by + cy + d = 0, $
 where $\{r\}$ and $\{a, b, c, d\}$ are the model parameters.

  \textbf{Intra-prediction Module.}  An example of the intra-prediction result of the plane modeling and the point modeling method for a car on the KITTI dataset is shown in Fig. \ref{fig:modeling_exp}. 
  We can predict the whole point cloud with the segmentation \textit{info. data} and the model parameters. 
 For a point in the predicted range image $\hat{\mathbf{I}}=(w, h, \hat{r})$ and its model parameters, we want to predict $\hat{r}$. Its 3D location $\hat{\mathbf{P}}=(\cos \varphi  \cdot \cos \theta  \cdot \hat{r}, \cos \varphi \cdot \sin \theta \cdot \hat{r}, \sin \varphi \cdot \hat{r})$, where the calculation of $\varphi$ and $\theta$ is in Sec. \ref{sec:tranform}. When the point is modeled by a point model $\{r\}$, then $\hat{r} = r$; if the point is modeled by a plane model $\{a, b, c, d\}$, and from the plane model equation, we can calculate the predicted $\hat{r}$:
 \begin{equation}
     \hat{r} = -d / (a \cdot \cos \varphi \cdot \cos \theta + b \cdot \cos \varphi \cdot \sin \theta + c \cdot \sin \varphi).
 \end{equation}

 \textbf{Key Point Extractor and Salience Map.}
  Since the point cloud is the front end in the autonomous system, the performance of downstream tasks relies on the reconstruction quality. But most downstream tasks only need the salient region of the point cloud to be accurate enough, not the whole point cloud. Thus, we propose a non-uniform compression and decompression framework to improve the compression efficiency to maintain the high performance of the downstream tasks.
  
  We select feature points that are on sharp edges and planar surface patches from LiDARs' raw measurements. We follow \cite{zhang2014loam} to evaluate the curvature of a point in a local region to classify it as an edge (high curvature) and planar point (low curvature) according to
  \begin{equation}
    c(\mathbf{P}_{i}) = \frac{||\sum_{\mathbf{P_{j}}\in \mathcal{S}, \mathbf{P}_{i}\neq\mathbf{P}_{j}}
    (\mathbf{P}_{i} - \mathbf{P}_{j})||}{|\mathcal{S}|\cdot ||\mathbf{P}_{i}||},
  \end{equation}
  where $\mathcal{S}$ is the set of consecutive points of $\mathbf{P}_{i}$ in the same scan.
  And then we set a series of salience score thresholds, to classify each cluster into different salience levels. For the clusters which have fewer key points, we will reduce the quantization accuracy of the residual data.
  
  
\textbf{Residual Quantization Module.}
The residual data are calculated by the range image minus the intra-predicted range image. For example in Fig. \ref{fig:modeling_exp}, the residual of point A: $res_{A} = r_A - \hat{r}_{A} = |OA| - |OA'|$. For the uniform compression framework, the residual data will be quantized into integers using uniform accuracy, and for the non-uniform framework, the residual data in the unimportant clusters (such as leaves on trees, or meadows) will be quantized into integers with lower quantization accuracy.
The maximum reconstruction error of each point is half the quantization accuracy.

\subsection{Decompression}
In the decompression module, the segmentation \textit{info. data} and the quantized residual data will be decoded from the same basic compressoras  in the compression. And the intra-prediction results can be obtained by the intra-prediction module. Then the dequantization module recovers the integer residual data into float numbers as the inverse process of the quantization module. Then we can reconstruct the whole range image by adding the residual data into the intra-predicted range image, which can be transformed into a reconstructed point cloud.

\section{Experiments}
\label{sec:exp}
\subsection{Evaluation Metrics}
In the compression stage, the original point cloud can be encoded into a bit stream for storage or transmission. The bit-per-point (BPP) and compression ratio (CR) are two naive metrics to evaluate the size of the compressed bitstream. In this paper, we only consider the geometry compression of the point cloud, so, the BPP of the original point cloud is 96 for the three float values x, y, and z. 

The reconstruction quality is evaluated by three evaluation metrics: F$_1$ score, point-to-point chamfer distance (CD) \cite{fan2017point,huang20193d}, and point-to-plane PSNR (D2 PSNR) \cite{mekuria2017performance,tian2017geometric}. For the original point cloud $P$ and reconstructed point cloud $\hat{P}$:

\begin{align}
    F_1 &= 2TP / (2TP + FP + FN),\\
    CD  &= CD_{sym}(P, \hat{P}) = \big[CD(P, \hat{P}) + CD(\hat{P}, P)\big] / 2, \label{eq:cd}\\
    PSNR&= 10\log_{10} \big[3r^2 / \max\{MSE(P, \hat{P}), MSE(\hat{P}, P)\}\big],
\end{align}
where the distance threshold of the F$_1$ score $\tau_{geo}=2$cm and  the peak constant value of the point-to-plane PSNR $r=59.70$m, and the other detailed definitions in the equations are the same as in \cite{biswas2020muscle}.
For downstream tasks, the bounding box average precision (AP) \cite{everingham2010pascal, sun2020scalability} of the classes car, pedestrian, and cyclist are evaluated for 3D object detection, and in SLAM, the absolute trajectory error (ATE) and the relative pose error (RPE) \cite{zhang2018tutorial} for translation and rotation are evaluated. 


\subsection{Dataset}
To evaluate the compression ability and quality in the point clouds with different densities, we choose three datasets: HKUSTCampus, Oxford\cite{RobotCarDatasetIJRR}, and KITTI\cite{geiger2013vision}, which were collected by a Velodyne VLP-16, Velodyne HDL-32E, and Velodyne HDL-64E, respectively. The HKUSTCampus dataset was collected with a handheld Velodyne at HKUST. Because the point clouds are achieved directly from the range images in HKUSTCampus, the points in the point clouds and pixels in the range images can be completely assigned; but the point clouds in Oxford and KITTI were pre-processed before publication. In the comparative experiments on the downstream tasks, the results of the original point cloud from the dataset (Raw Data), and point cloud backprojected from a range image (Original Data) are compared both.

The point cloud distribution in different scenes can be very different. We choose city, residential, campus, and road in the KITTI raw data, like \cite{sun2019novel}, to evaluate the compression ratio in various scenes. The point clouds in the first three scenes are denser and better-regulated than in the road scene.

\subsection{Ablation Study}
\label{sec:ablation}
\begin{figure}[t]
\centering
\includegraphics[width=0.38\paperwidth]{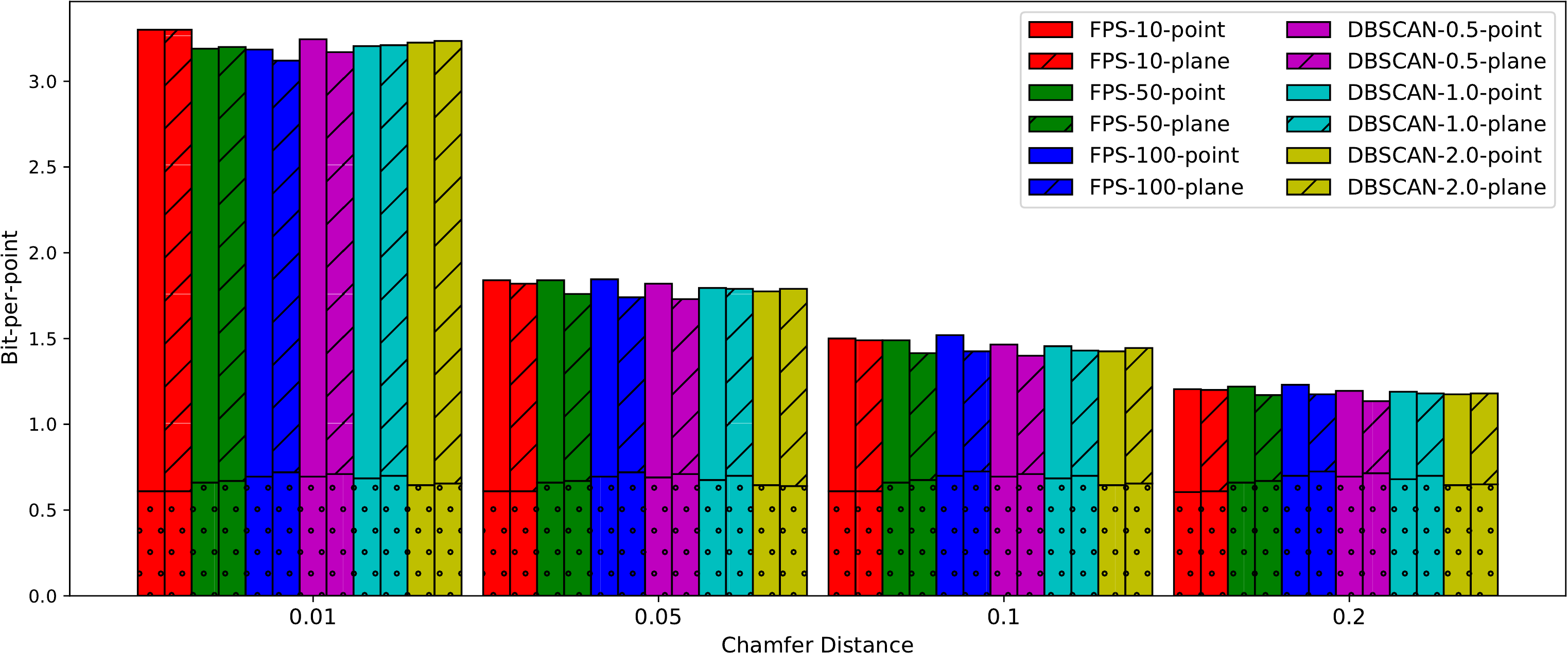}
\caption{The chart diagram of BPP with different methods. For example, \textit{FPS-10-point} means the segmentation method is FPS, cluster number is 10, and model method is \textit{point} only; \textit{DBSCAN-2.0-plane} means DBSCAN with neighbor distance (eps) 2.0m and the model method is \textit{plane/point}.}
\label{fig:ablation2_1}
\vspace{-0.1cm}
\end{figure}

\textbf{Segmentation and Modeling.} 
The segmentation methods control the cluster size and residual range, and the modeling methods vary the surf residual distribution. We implement \textit{open3D} DBSCAN as the instance-based segmentation baseline as in \cite{sun2019novel}. 
The comparative results of \textit{FPS-} and \textit{DBSCAN-} in Fig. \ref{fig:ablation2_1} shows that our FPS-based segmentation method is better than instance-based segmentation. Object-based segmentation is not necessary in the compression tasks when using BZip2 as basic compressor. Fig. \ref{fig:ablation2_1} also shows that the residual and residual BPP decrease when the cluster is smaller and denser. The overall BPP is best when the number in the cluster is from 50 to 100. 

Tab. \ref{tab:ablation2_2} shows that the plane modeling method in our framwork can largely decrease the mean of the residual data, especially in the KITTI city dataset, because there are more walls and planes in the city scene. In the KITTI road dataset, the point modeling method is better than the plane modeling method. 



\begin{table*}[tbp]
\renewcommand\arraystretch{1.1}   
\renewcommand\tabcolsep{8pt}  
\setlength{\abovecaptionskip}{0pt} 
\setlength{\belowcaptionskip}{0pt} 
\captionsetup{justification=centering}
	\caption{\textsc{Comparative Results of information BPP, residual BPP, and mean of residual in four different settings on four KITTI Datasets in scene city, residential, campus, and road. }}
	\label{tab:ablation2_2} \centering %
\begin{tabular}{cccccccllllll}
\hline
\multirow{2}{*}{}  & \multicolumn{3}{c}{FPS+point} & \multicolumn{3}{c}{FPS+plane} & \multicolumn{3}{c}{DBSCAN+point}                                              & \multicolumn{3}{c}{DBSCAN+plane}                                              \\ \cmidrule(l){2-4} \cmidrule(l){5-7}  \cmidrule(l){8-10}  \cmidrule(l){11-13} 
                   & IBPP     & RBPP     & Res     & IBPP     & RBPP     & Res     & \multicolumn{1}{c}{IBPP} & \multicolumn{1}{c}{RBPP} & \multicolumn{1}{c}{Res} & \multicolumn{1}{c}{IBPP} & \multicolumn{1}{c}{RBPP} & \multicolumn{1}{c}{Res} \\ \hline
KITTI\_city        & 0.6      & 1.8      & 0.7     & 0.65     & 1.74     & 0.56    & 0.56                     & 1.78                     & 1.68                    & 0.59                     & 1.76                     & 1.5                     \\
KITTI\_residential & 0.51     & 2.57     & 0.57    & 0.55     & 2.55     & 0.52    & 0.43                     & 2.64                     & 2.35                    & 0.44                     & 2.64                     & 2.22                    \\
KITTI\_campus      & 0.93     & 2.53     & 0.93    & 0.99     & 2.53     & 0.9     & 1.01                     & 2.54                     & 1.69                    & 1.08                     & 2.52                     & 1.67                    \\
KITTI\_road        & 1.08     & 4.05     & 0.8     & 1.11     & 4.09     & 0.81    & 1.03                     & 4.24                     & 2.39                    &       1.03                   & 4.26    & 2.38    \\ \hline
\end{tabular}
\vspace{-0.3cm}
\end{table*}

\textbf{Basic Compressor.}
In this section, we implement different basic compressors to show the compression ratios of our compression framework and the baseline point cloud compression. In the baseline method, the single basic compressor is used to encode the quantized point cloud with the same quantization accuracy as the residual quantization in our method. In our method with different basic compressor, the number of clusters is set to 100 and the segmentation and modeling method is \textit{FPS+plane}. 

\begin{figure}[t]
\centering
\includegraphics[width=0.4\paperwidth]{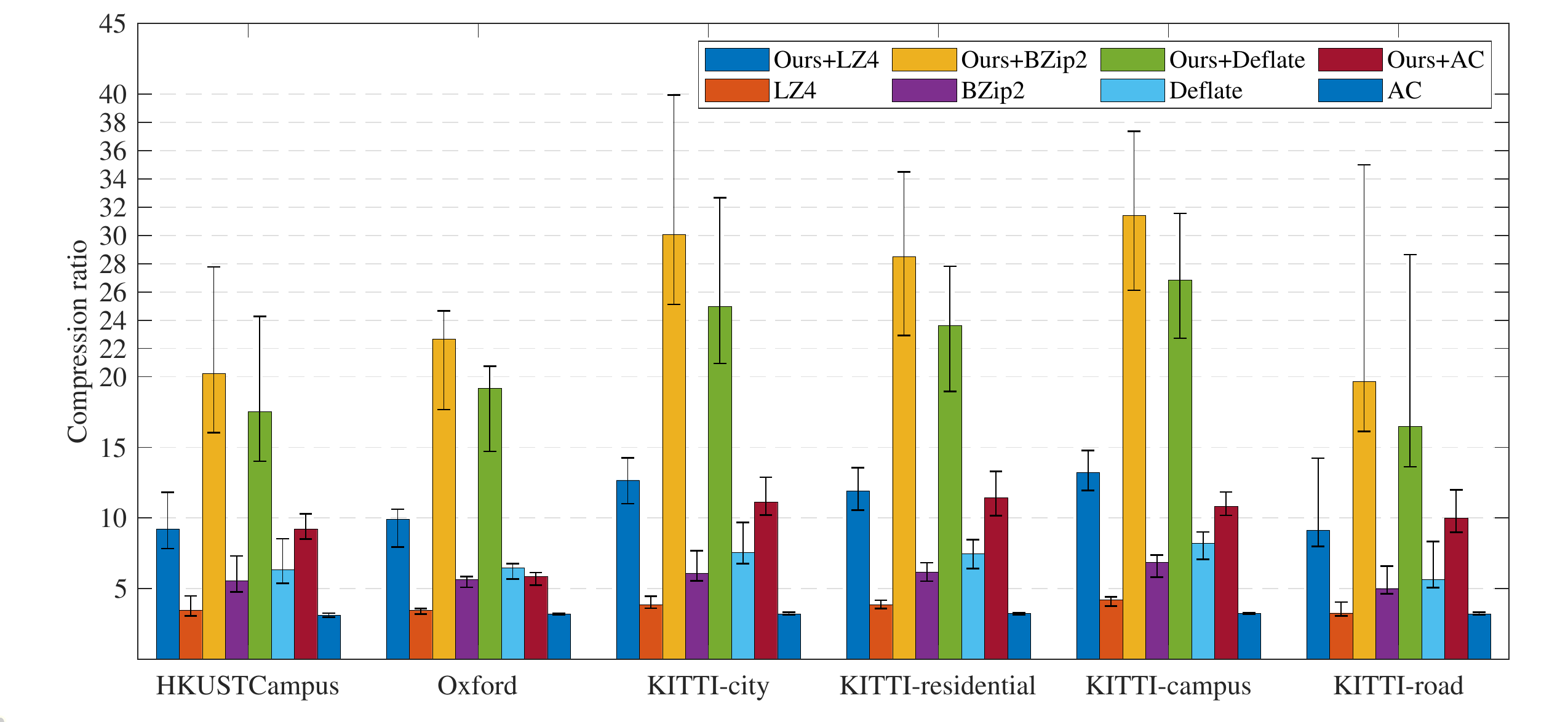}
\caption{The chart diagram of the compression ratio of our method with different basic compressors. Three datasets, HKUSTCampus, Oxford \cite{RobotCarDatasetIJRR}, and different scenes in KITTI \cite{geiger2013vision} are tested. Single basic compressors are directly applied to compress the quantized point cloud as the baseline method.}
\label{fig:ablation1}
\vspace{-0.1cm}
\end{figure}

Fig. \ref{fig:ablation1} shows the results of the compression ratio with the different scenes and basic compressors. Comparing the results of different scenes, the conclusion is that the denser of the point cloud, the higher of the compression ratio. From Fig. \ref{fig:ablation1} and Tab. \ref{tab:time_basic_compressor} we can find that the compression performance of BZip2 is the best in our experiments, but the speed is also the slowest.

\begin{table}[tbp]
\renewcommand\arraystretch{1.1}   
\renewcommand\tabcolsep{4.5pt}  
\setlength{\abovecaptionskip}{0pt} 
\setlength{\belowcaptionskip}{0pt} 
\captionsetup{justification=centering}
	\caption{\textsc{The encoding and decoding time cost of different compressors.}}
	\label{tab:time_basic_compressor} \centering %
\begin{tabular}{ccccccccc}
\hline
\multirow{2}{*}{} & \multicolumn{4}{c}{Encoding time (ms)}     & \multicolumn{4}{c}{Decoding time (ms)}                                    \\
 \cmidrule(l){2-5}  \cmidrule(l){6-9} 
                  & LZ4  & BZip2 & Deflate & AC    & LZ4           & BZip2         & Deflate       & AC            \\ \hline
16E               & 0.03 & 11.4  & 3.7     & 40    & \textbf{0.03} & \textbf{5.2}  & \textbf{1.3}  & \textbf{19.1} \\
32E               & 0.04 & 26.5  & 9.7     & 92    & \textbf{0.03} & \textbf{11.7} & \textbf{3.72} & \textbf{43.4} \\
64E               & 0.1  & 47.9  & 17.5    & 161.7 & \textbf{0.1}  & \textbf{20.9} & \textbf{6.8}  & \textbf{75.5} \\ \hline
\end{tabular}
\vspace{-0.3cm}
\end{table}

\begin{figure}[t]
\centering
\begin{minipage}[t]{0.32\linewidth}
\centering
\includegraphics[width=1.0\textwidth]{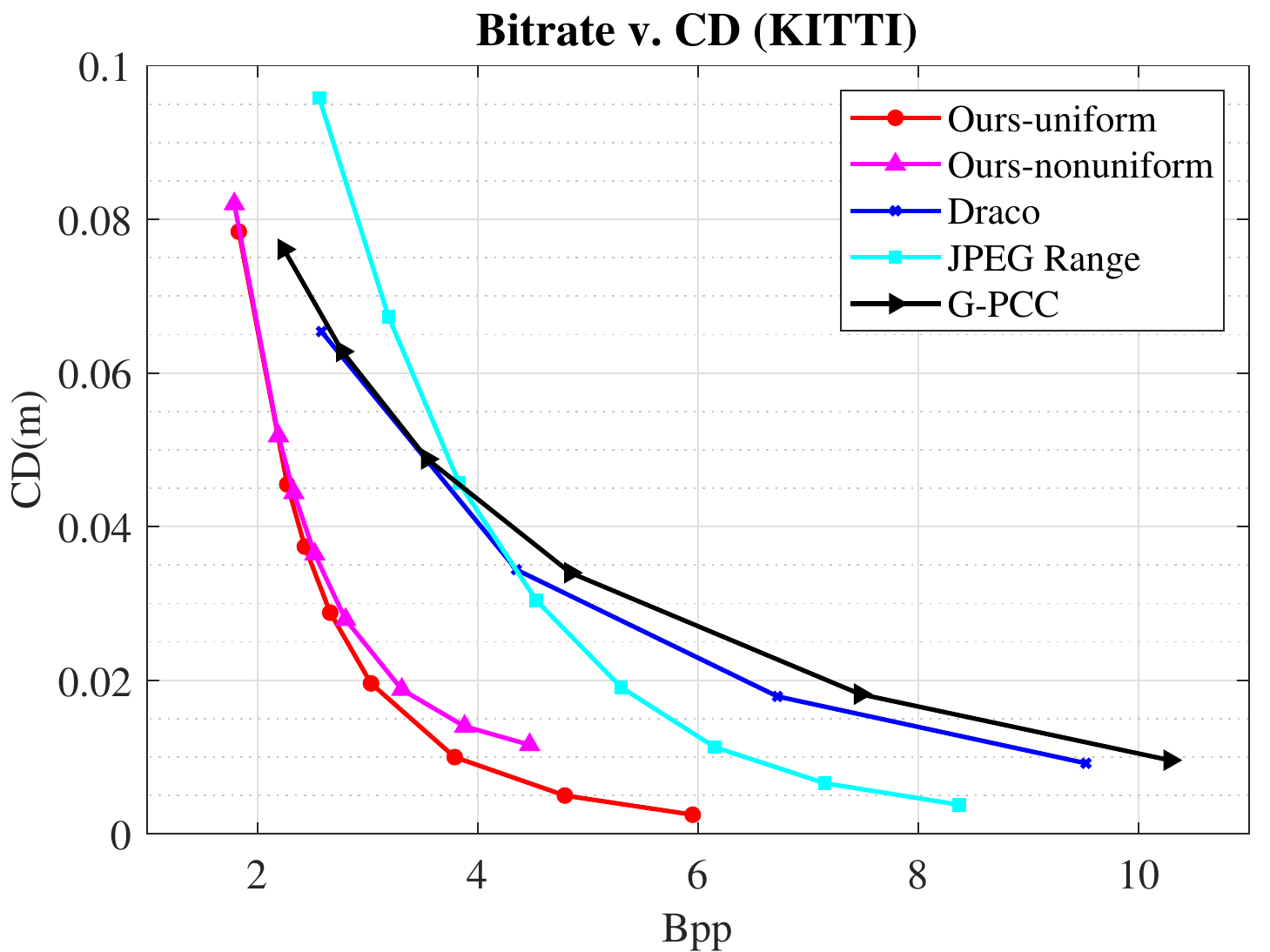}
\end{minipage}
\hfill
\begin{minipage}[t]{0.32\linewidth}
\centering
\includegraphics[width=1.0\textwidth]{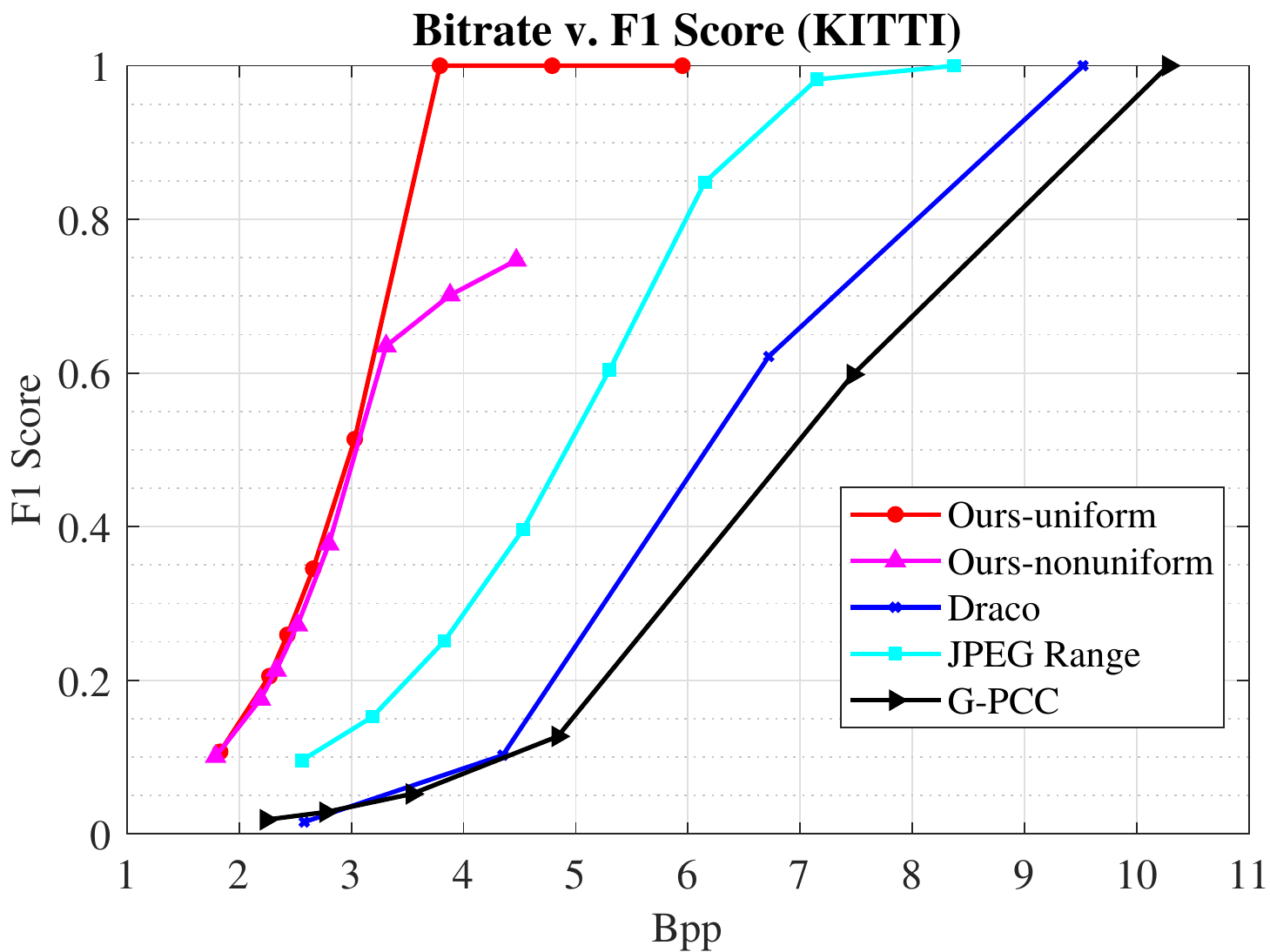}
\end{minipage}
\hfill
\begin{minipage}[t]{0.32\linewidth}
\centering
\includegraphics[width=1.0\textwidth]{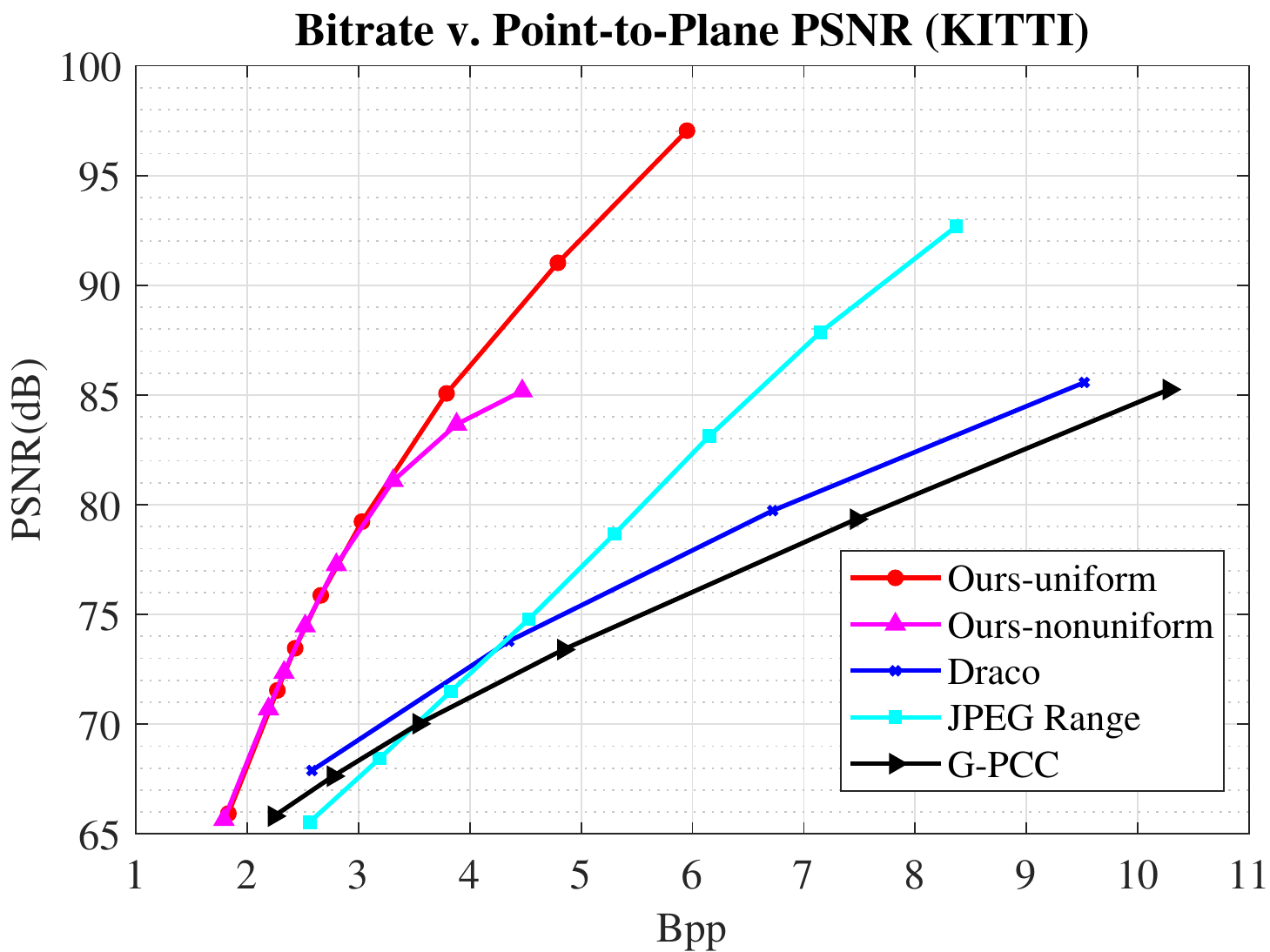}
\end{minipage}
\caption{Quantitative results on KITTI city dataset. Bit-per-point \textit{vs}. symmetric chamfer distance ($\downarrow$), F1 score (with $\tau_{geo}=0.02$m) ($\uparrow$), and point-to-plane PSNR (with $r=59.70$) ($\uparrow$) are shown from left to right. }
\label{fig:compare_1}
\vspace{-0.4cm}
\end{figure}

\subsection{Comparative Results}
In this section, we compare our proposed uniform and non-uniform compression frameworks with the baseline point cloud compression methods: the KDTree-based algorithm from Google: Draco\cite{GoogleDraco}; geometry-based compression method: G-PCC\cite{schwarz2018emerging, MPEG}; and range image-based compression using image compression method JPEG2000 (JPEG Range) \cite{wallace1992jpeg, rabbani2002jpeg2000}. Our compression frameworks use \textit{FPS+plane} as segmentation and modeling method, the number of clusters is 100, the basic compressor is BZip2, and the dataset is KITTI city. More specifically, our non-uniform compression method classifies the clusters into four saliency levels, and the rules for the classification and non-uniform quantization accuracy are shown in Tab. \ref{tab:salient_level}. For example, if the basic accuracy is 0.02m, and one cluster has 7 key points, then the saliency level of this cluster is 1 and the quantization accuracy of the residual data in this cluster is $0.02 + 0.04=0.06$m.
We compare the quality of the reconstructed point cloud and the performance of different downstream tasks in this section. 

\begin{table}[tbp]
\vspace{-0.4cm}
\renewcommand\arraystretch{1.1}   
\renewcommand\tabcolsep{5pt}  
\setlength{\abovecaptionskip}{0pt} 
\setlength{\belowcaptionskip}{0pt} 
\captionsetup{justification=centering}
	\caption{\textsc{Clusters' saliency level with number of key points and quantization accuracy.}}
	\label{tab:salient_level} \centering %
\begin{tabular}{ccccc}
\hline
                 & level 0  & level 1   & level 2    & level 3     \\ \hline
Key Point Num    & {[}0, 3) & {[}3, 10) & {[}10, 30) & {[}30, inf) \\
Quantization Acc & +0.06    & +0.04     & +0.02      & Base Acc    \\ \hline
\end{tabular}
\vspace{-0.3cm}
\end{table}

\begin{figure*}[th]
\centering
\begin{minipage}[t]{0.31\linewidth}
\centering
\includegraphics[width=1.0\textwidth]{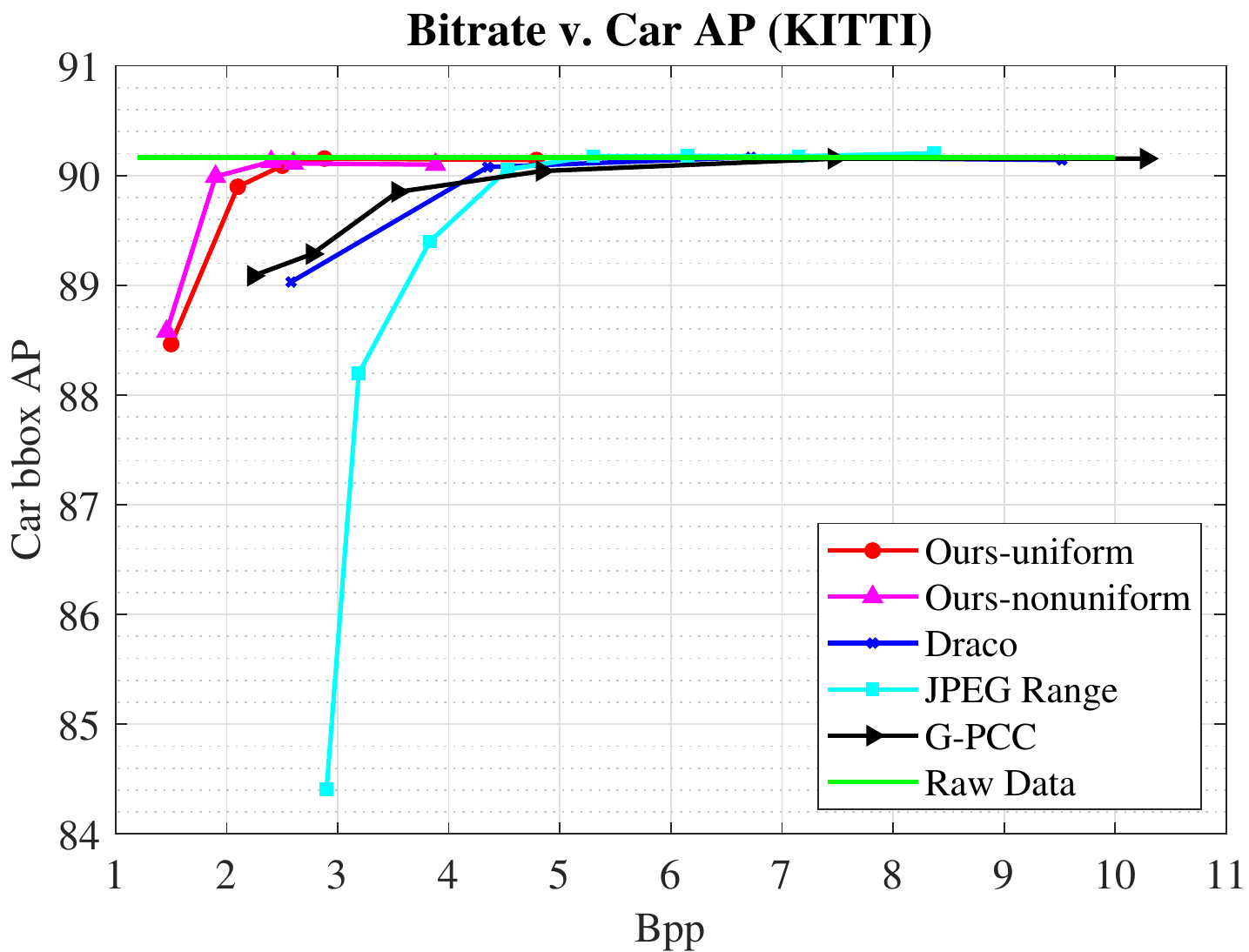}
\end{minipage}
\hfill
\begin{minipage}[t]{0.31\linewidth}
\centering
\includegraphics[width=1.0\textwidth]{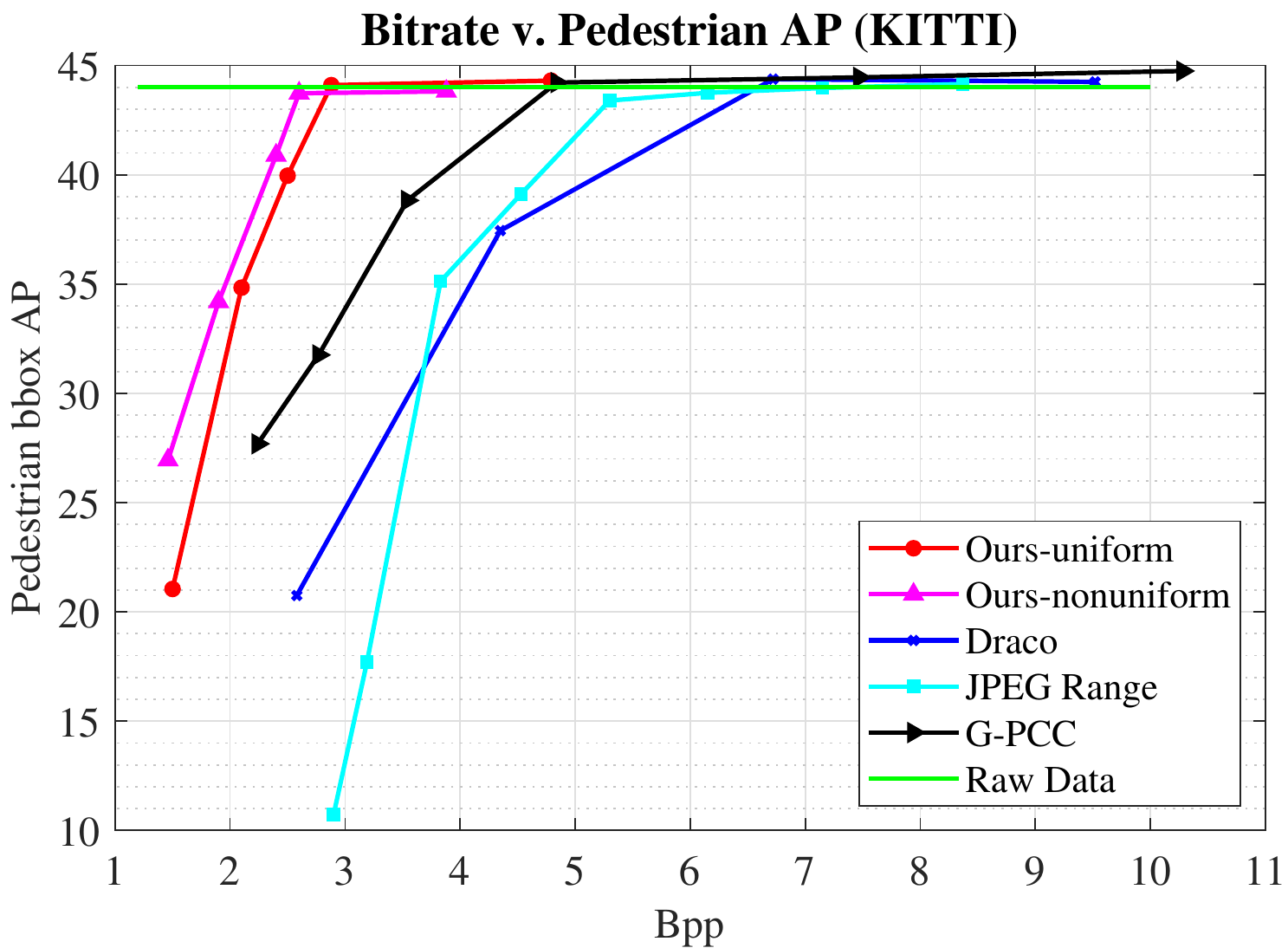}
\end{minipage}
\hfill
\begin{minipage}[t]{0.31\linewidth}
\centering
\includegraphics[width=1.0\textwidth]{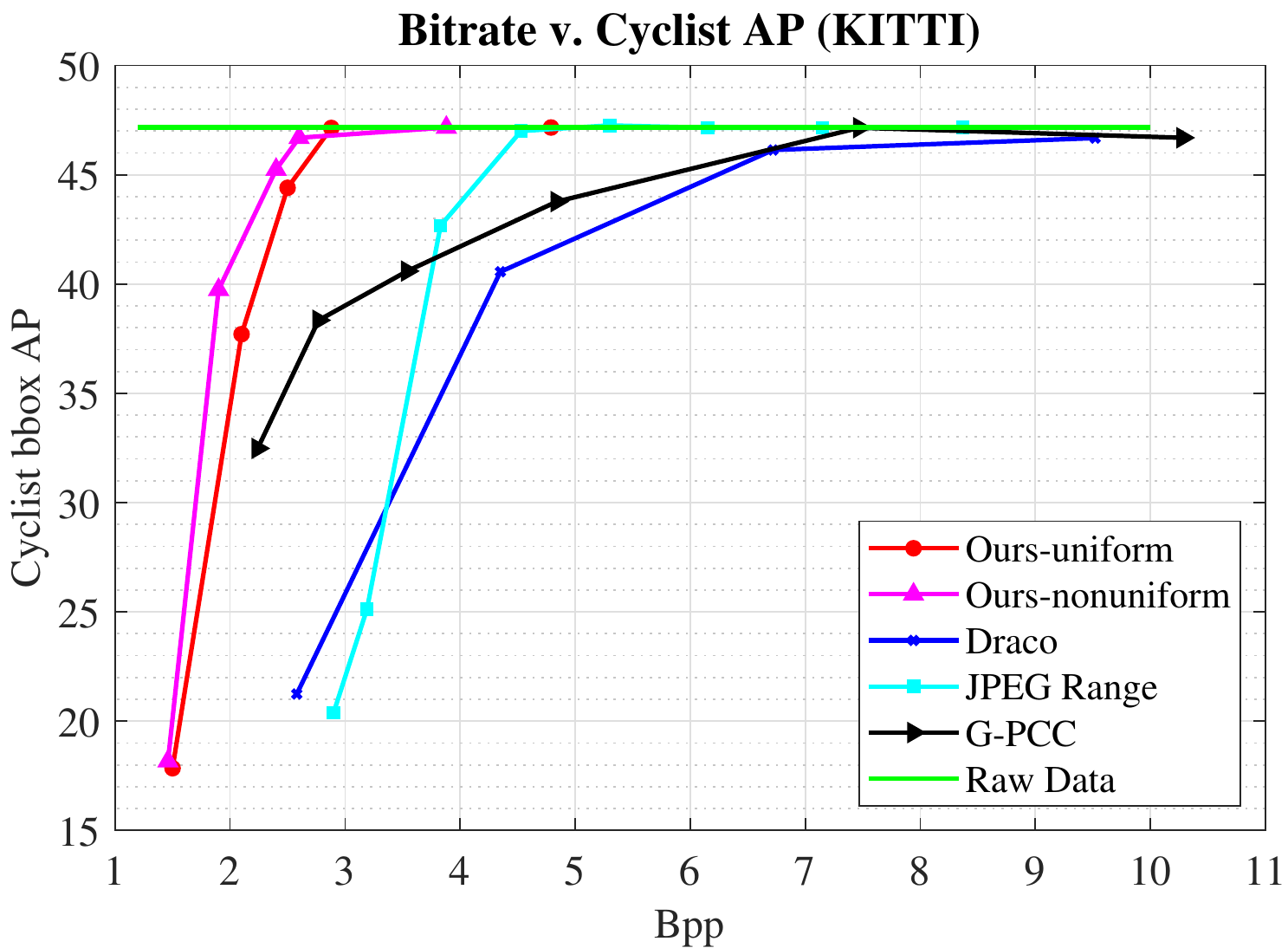}
\end{minipage}
\vfill
\centering
\begin{minipage}[t]{0.24\linewidth}
\centering
\includegraphics[width=1.0\textwidth]{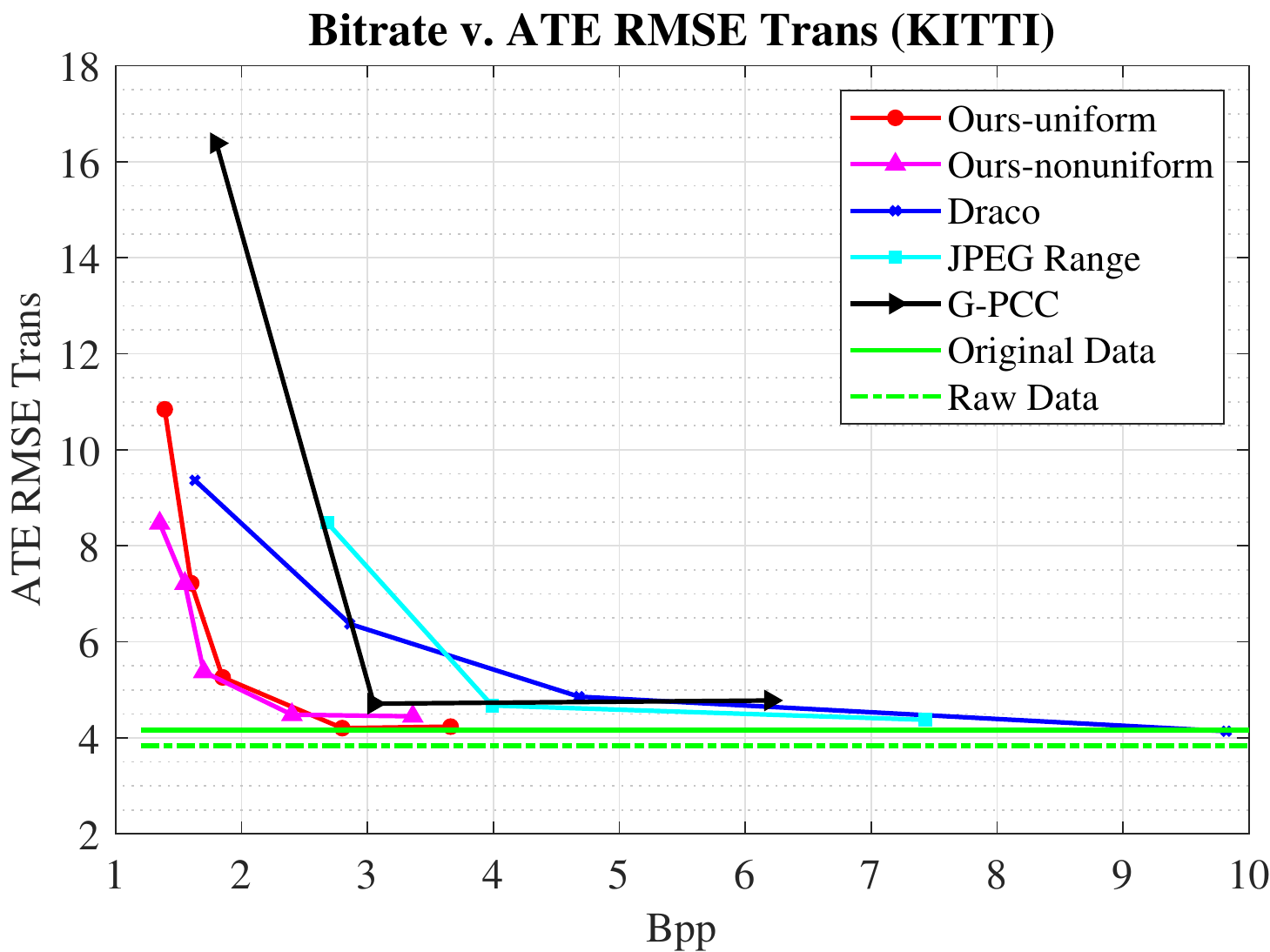}
\end{minipage}
\hfill
\begin{minipage}[t]{0.24\linewidth}
\centering
\includegraphics[width=1.0\textwidth]{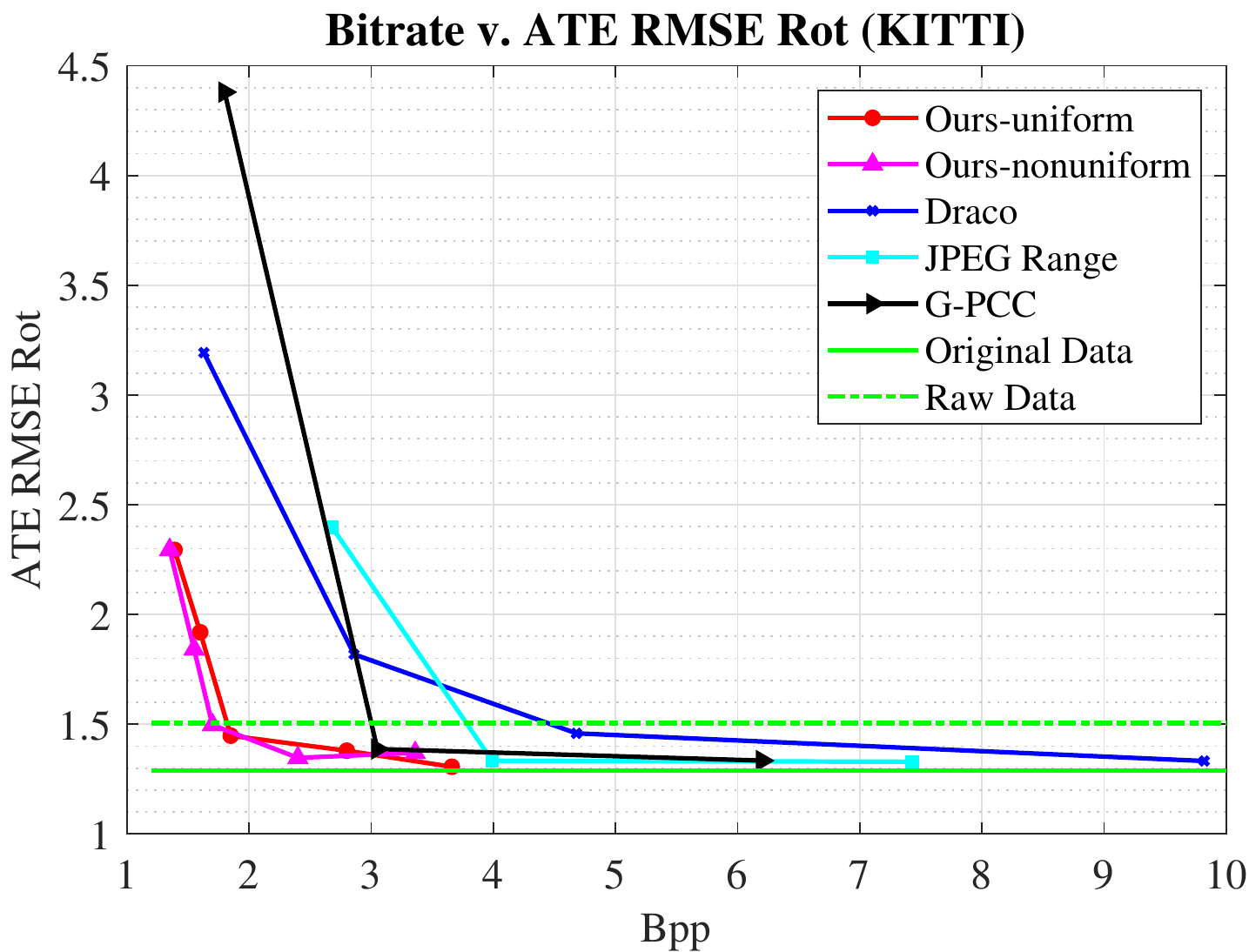}
\end{minipage}
\hfill
\begin{minipage}[t]{0.24\linewidth}
\centering
\includegraphics[width=1.0\textwidth]{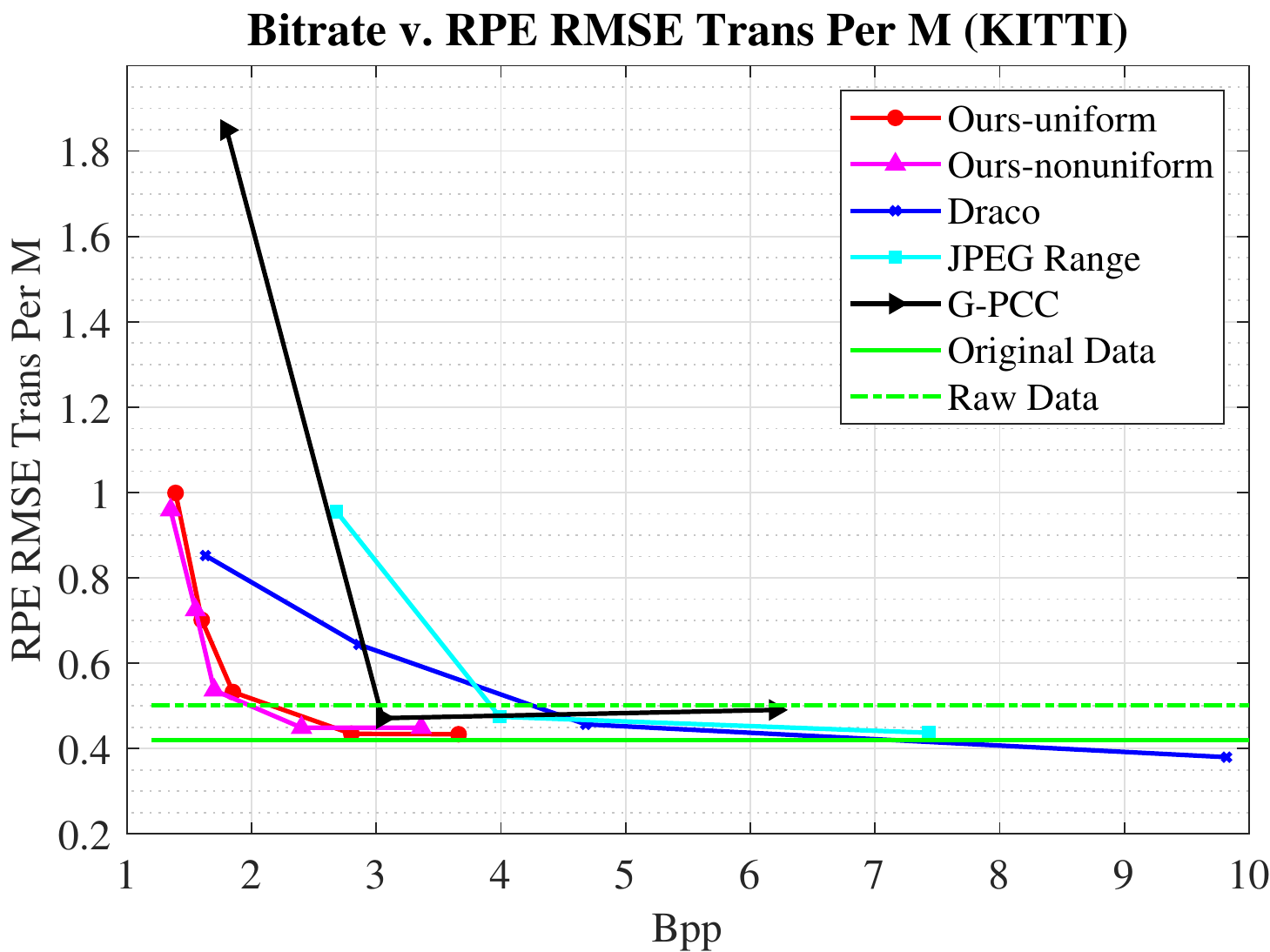}
\end{minipage}
\hfill
\begin{minipage}[t]{0.24\linewidth}
\centering
\includegraphics[width=1.0\textwidth]{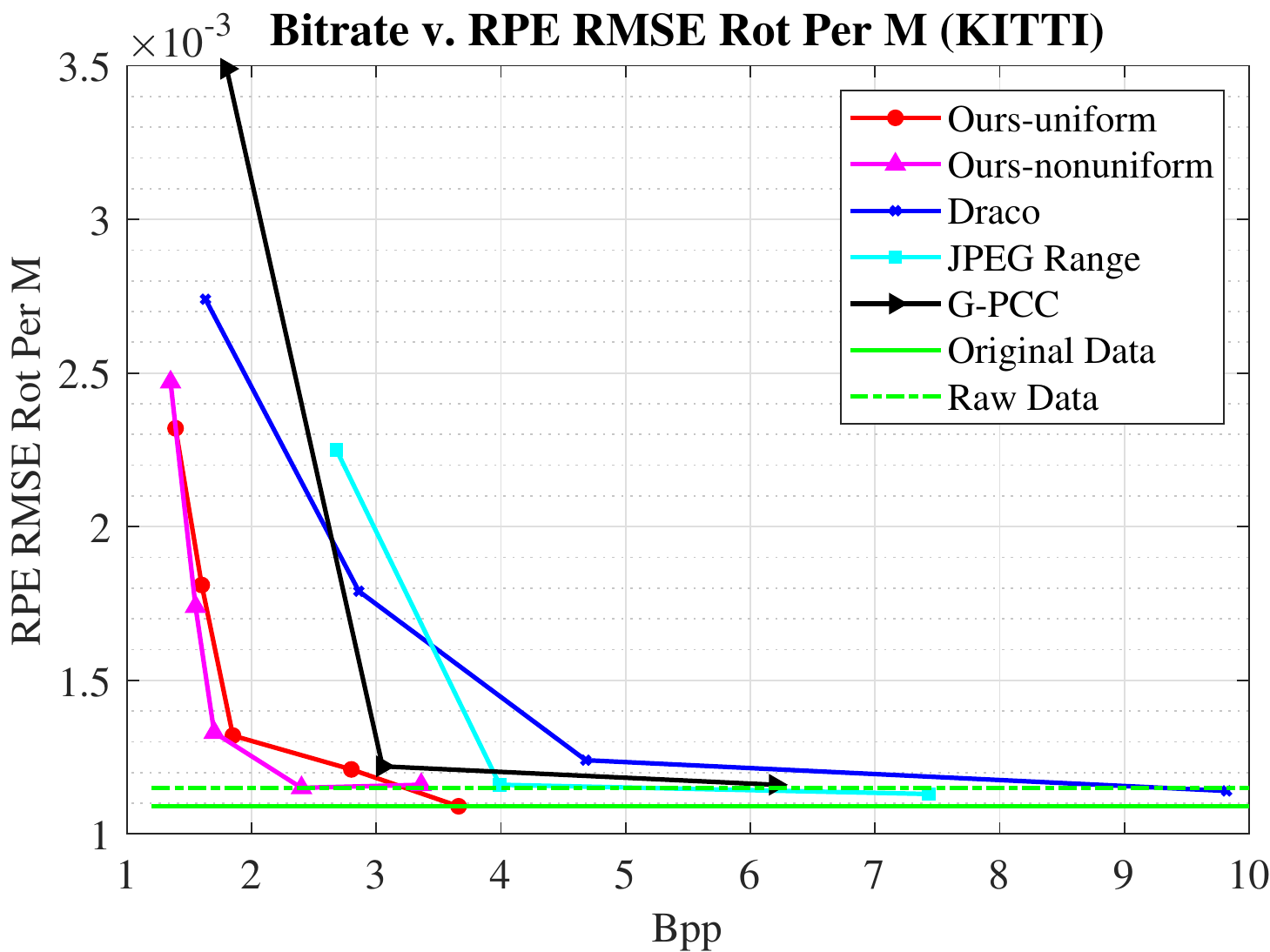}
\end{minipage}
\caption{Quantitative results of the 3D object detection (PointPillar) on the KITTI detection dataset and SLAM (A-LOAM) on the KITTI odometry dataset (seq 00). Bit-per-point \textit{vs}. Car BBox AP@0.7, 0.7, 0.7 ($\uparrow$),  Pedestrian BBox AP@0.5, 0.5, 0.5 ($\uparrow$), and  Cyclist BBox AP@0.5, 0.5, 0.5 ($\uparrow$), ATE RMSE Trans ($\downarrow$), ATE RMSE Rot ($\downarrow$), RPE RMSE Trans Per M ($\downarrow$), and RPE RMSE Rot Per M ($\downarrow$) are shown from top to bottom, left to right. }
\label{fig:compare_2}
\vspace{-0.1cm}
\end{figure*}

\begin{figure*}[th]
\centering
\begin{minipage}[t]{0.19\linewidth}
\centering
\includegraphics[width=1.0\linewidth]{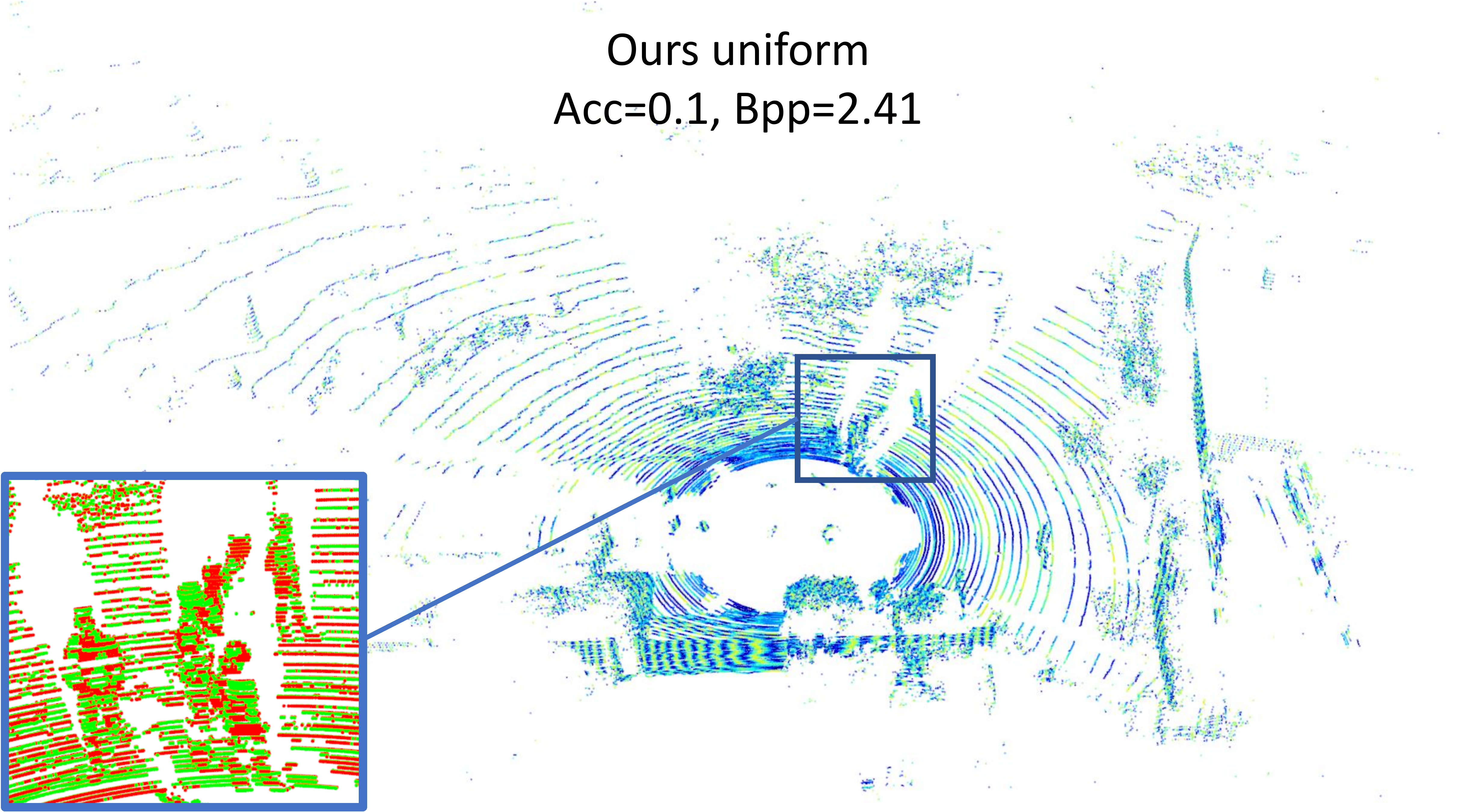}
\end{minipage}
\hfill
\centering
\begin{minipage}[t]{0.19\linewidth}
\centering
\includegraphics[width=1.0\linewidth]{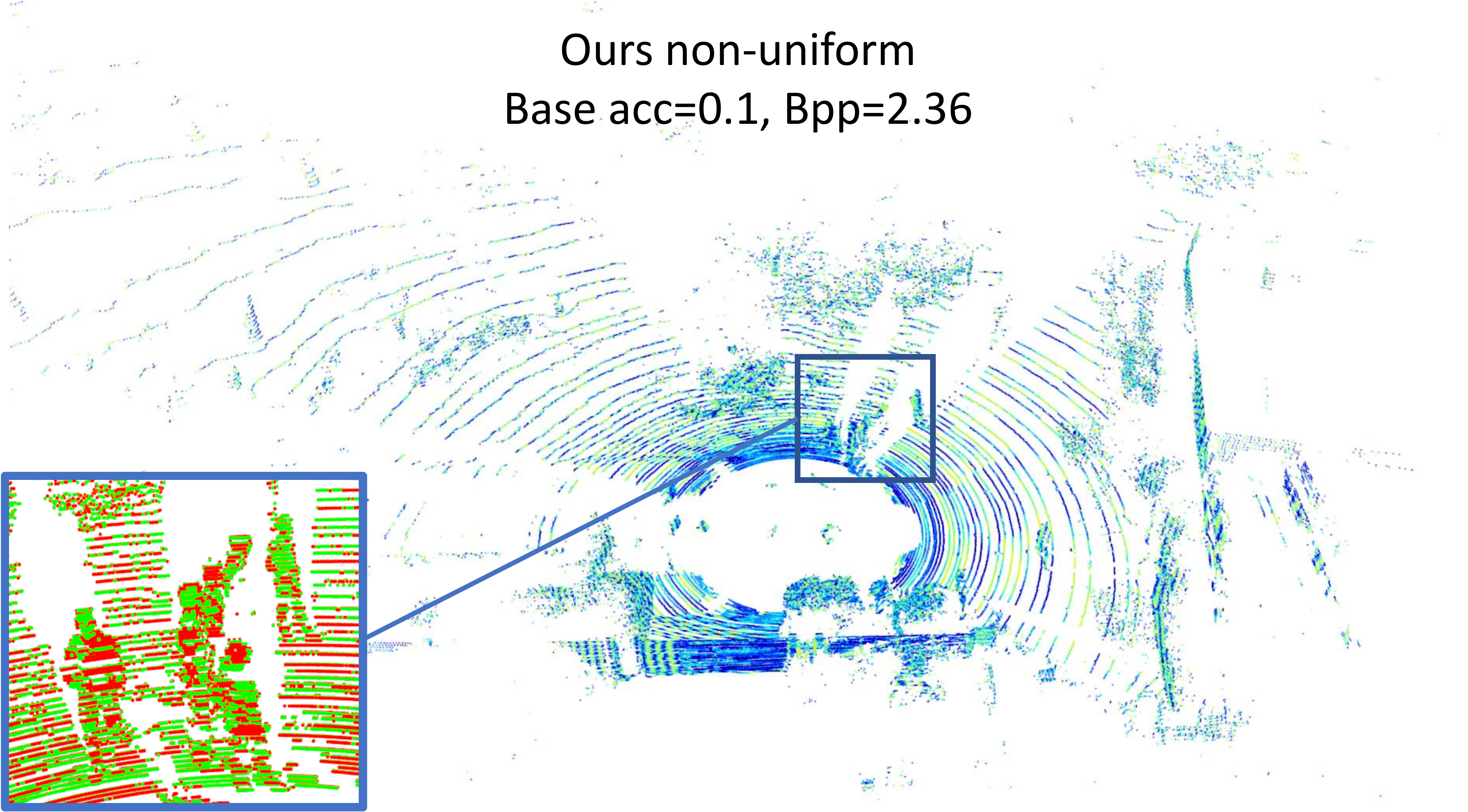}
\end{minipage}
\hfill
\centering
\begin{minipage}[t]{0.19\linewidth}
\centering
\includegraphics[width=1.0\linewidth]{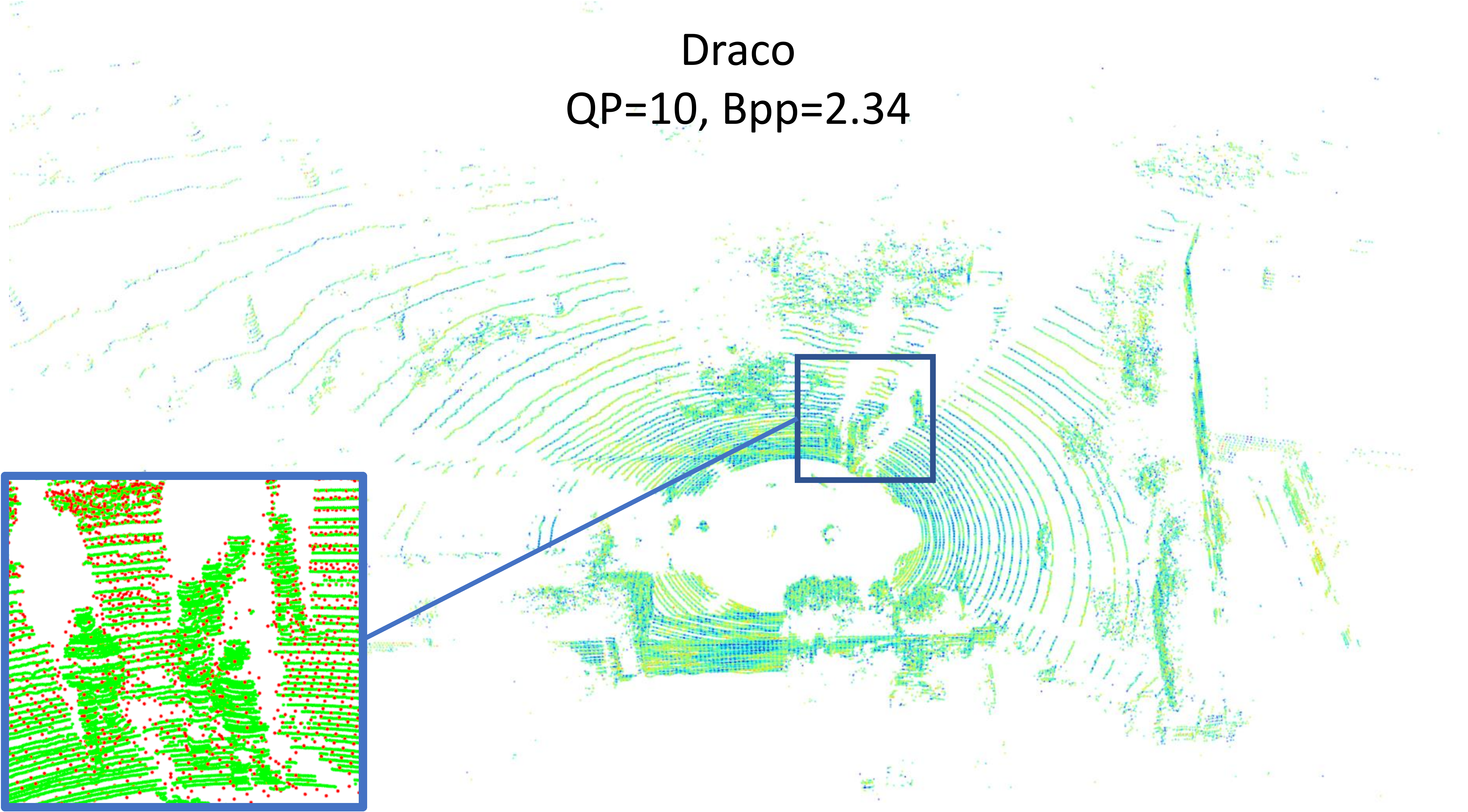}
\end{minipage}
\hfill
\centering
\begin{minipage}[t]{0.19\linewidth}
\centering
\includegraphics[width=1.0\linewidth]{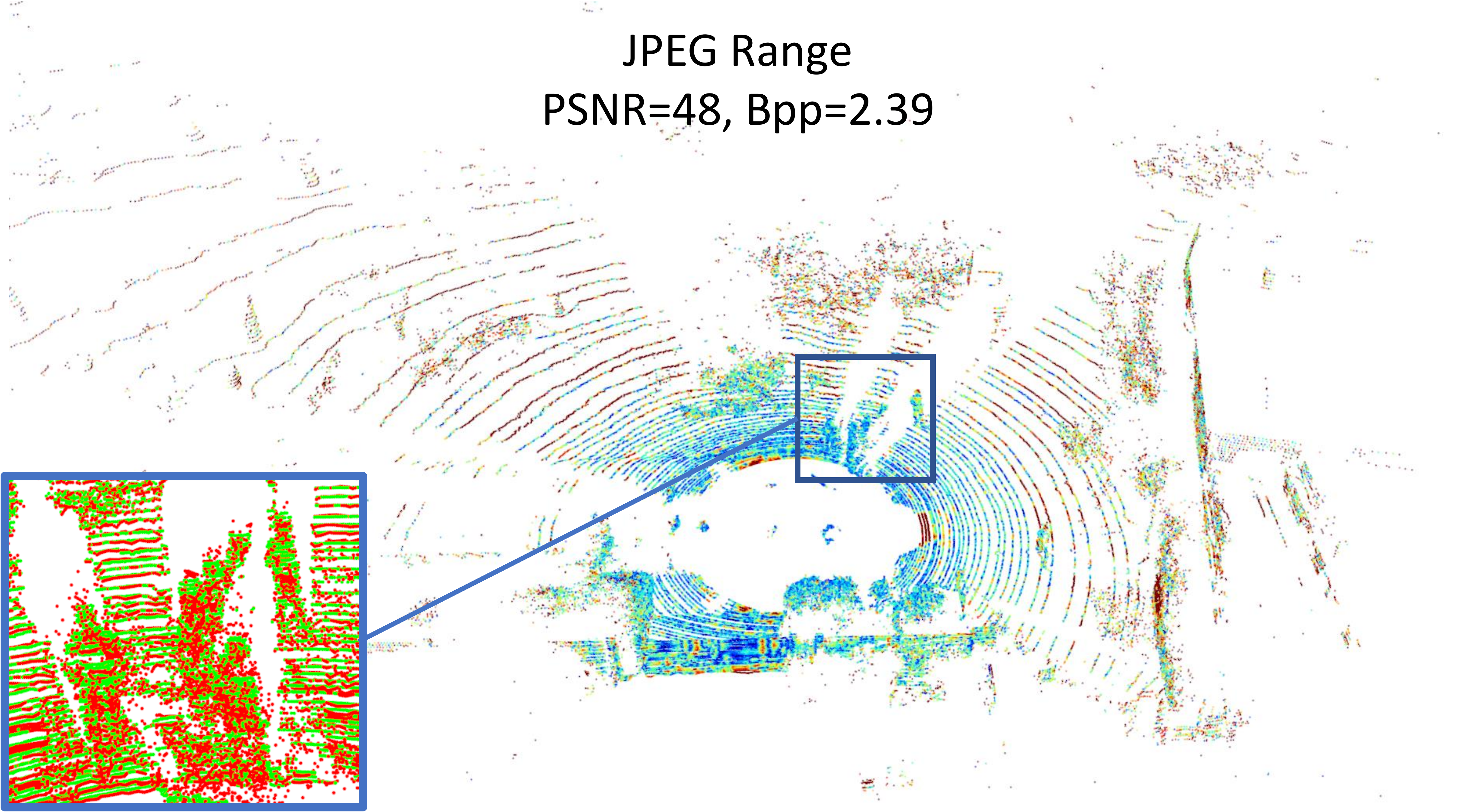}
\end{minipage}
\hfill
\centering
\begin{minipage}[t]{0.19\linewidth}
\centering
\includegraphics[width=1.0\linewidth]{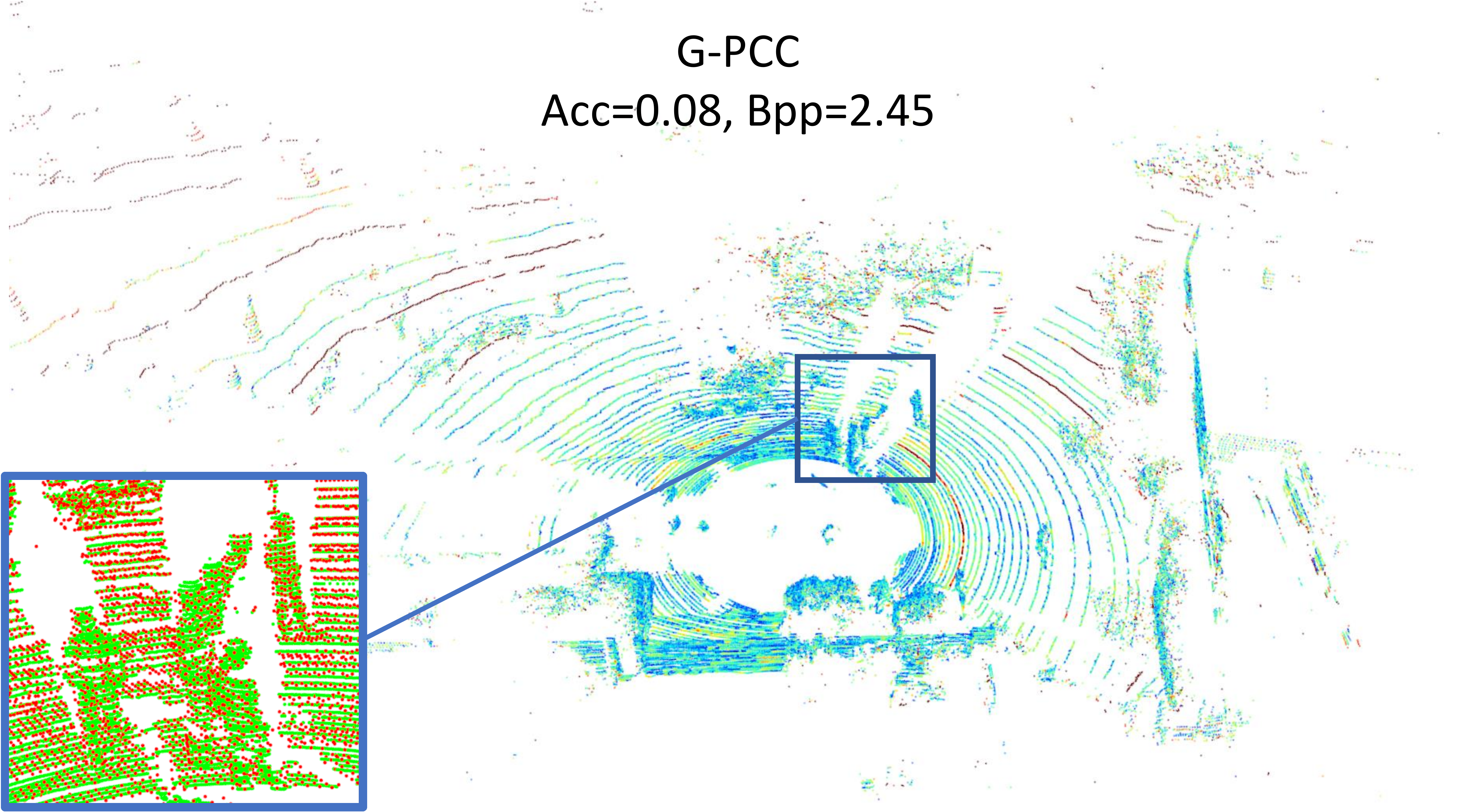}
\end{minipage}
\vspace{0.2cm}
\vfill
\begin{minipage}[t]{0.5\linewidth}
\centering
\includegraphics[width=1.0\textwidth]{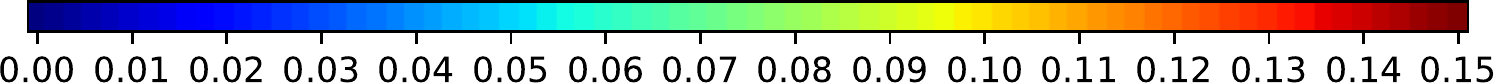}
\end{minipage}
\vspace{0.2cm}
\vfill
\centering
\begin{minipage}[t]{0.19\linewidth}
\centering
\includegraphics[width=1.0\linewidth]{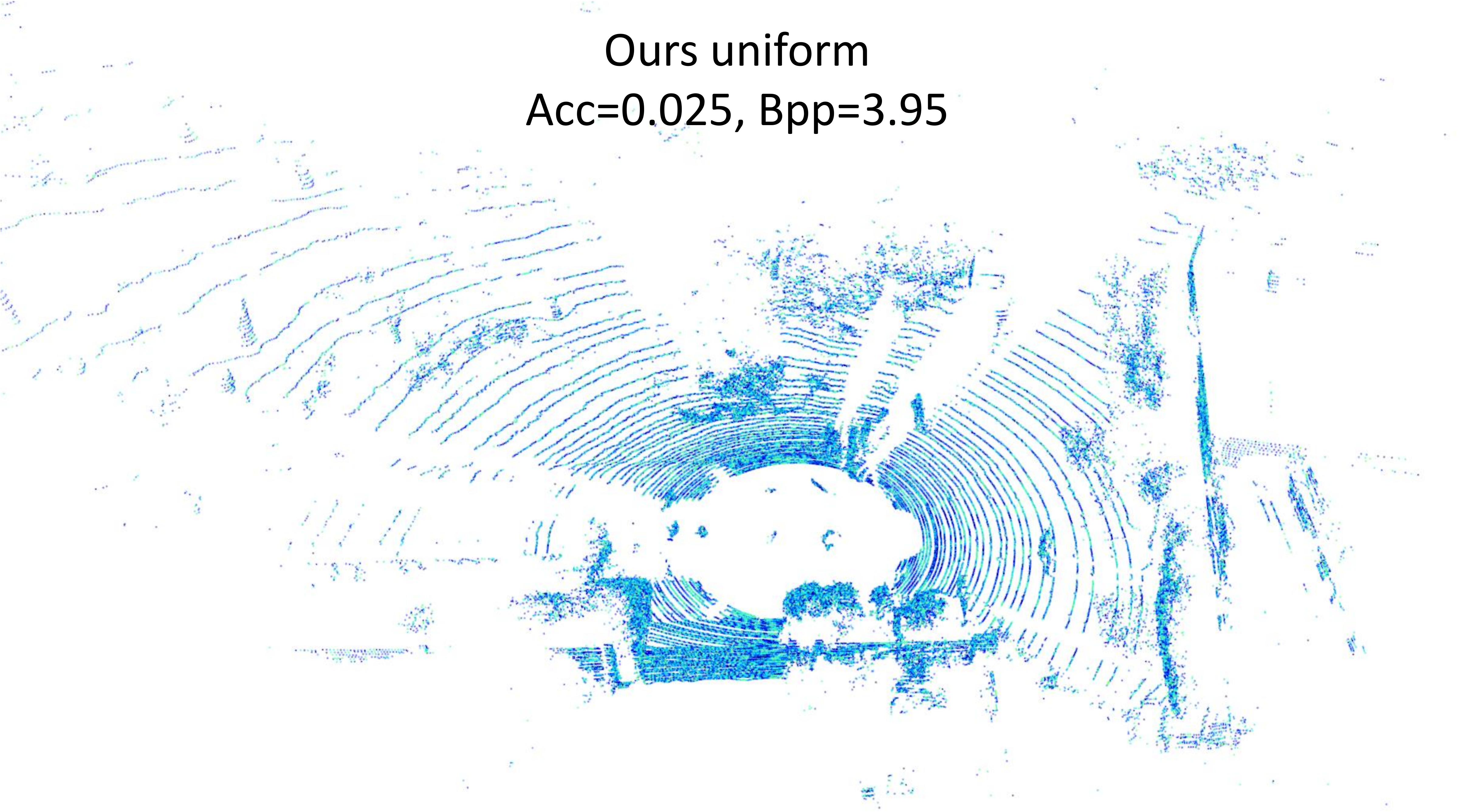}
\end{minipage}
\hfill
\centering
\begin{minipage}[t]{0.19\linewidth}
\centering
\includegraphics[width=1.0\linewidth]{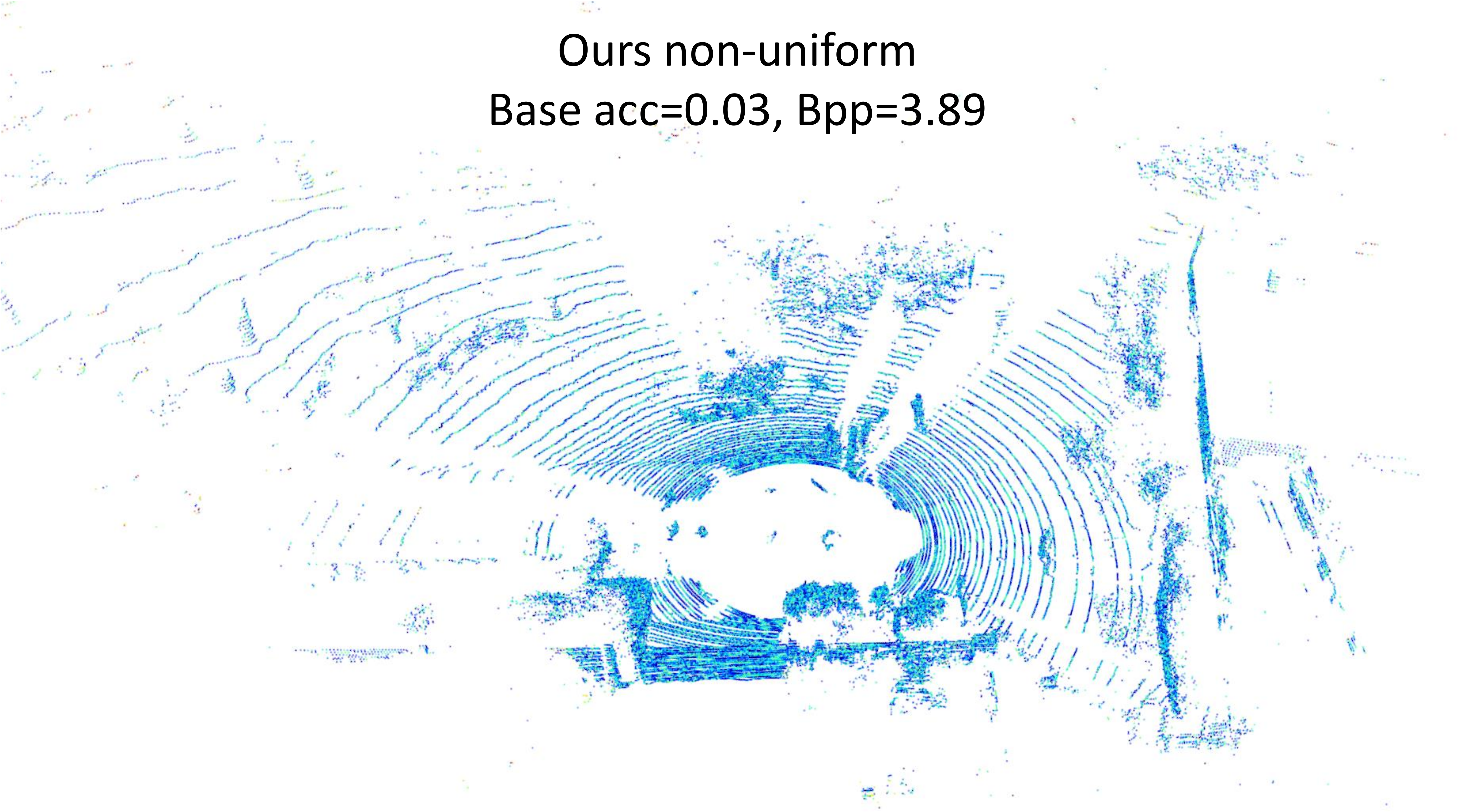}
\end{minipage}
\hfill
\centering
\begin{minipage}[t]{0.19\linewidth}
\centering
\includegraphics[width=1.0\linewidth]{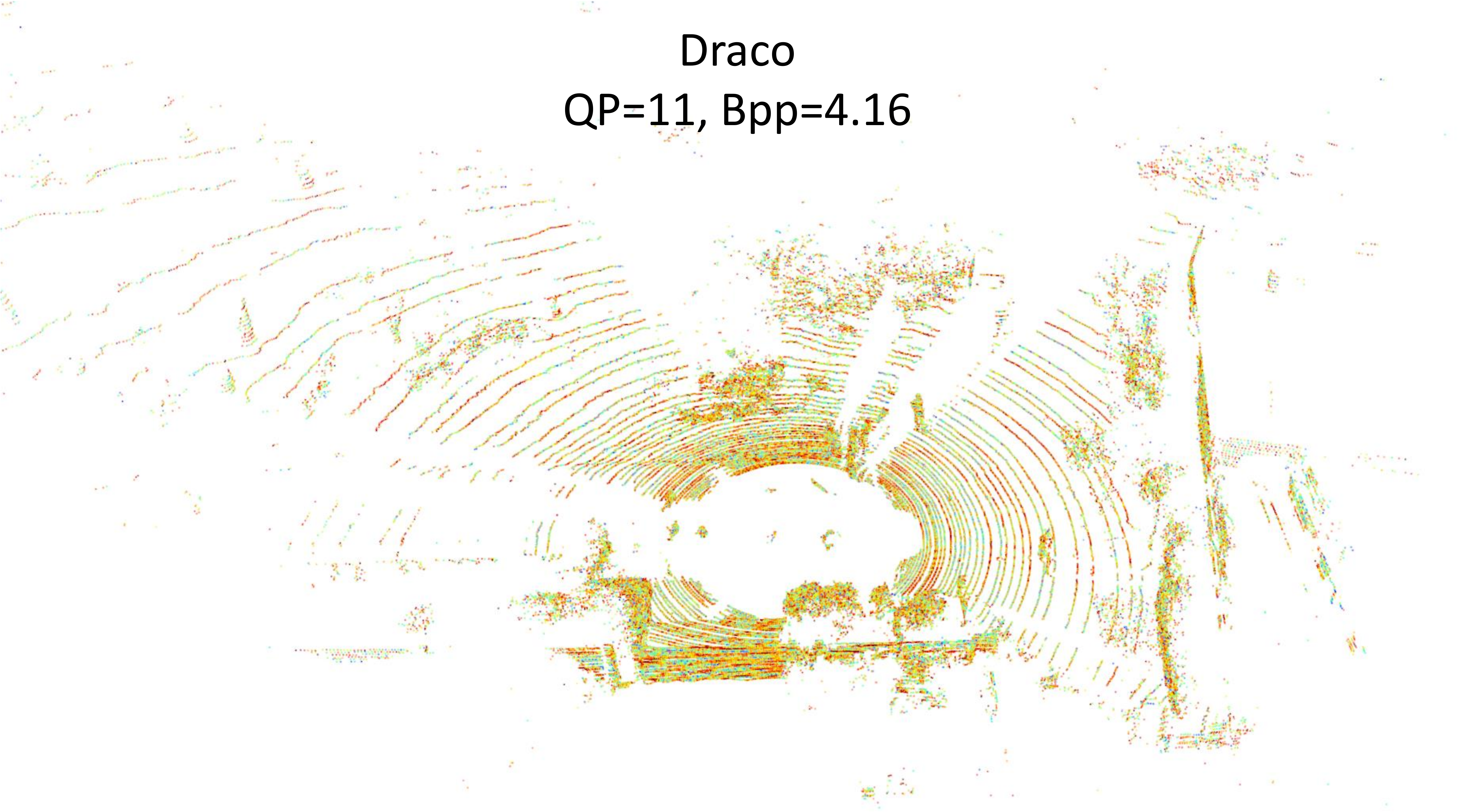}
\end{minipage}
\hfill
\centering
\begin{minipage}[t]{0.19\linewidth}
\centering
\includegraphics[width=1.0\linewidth]{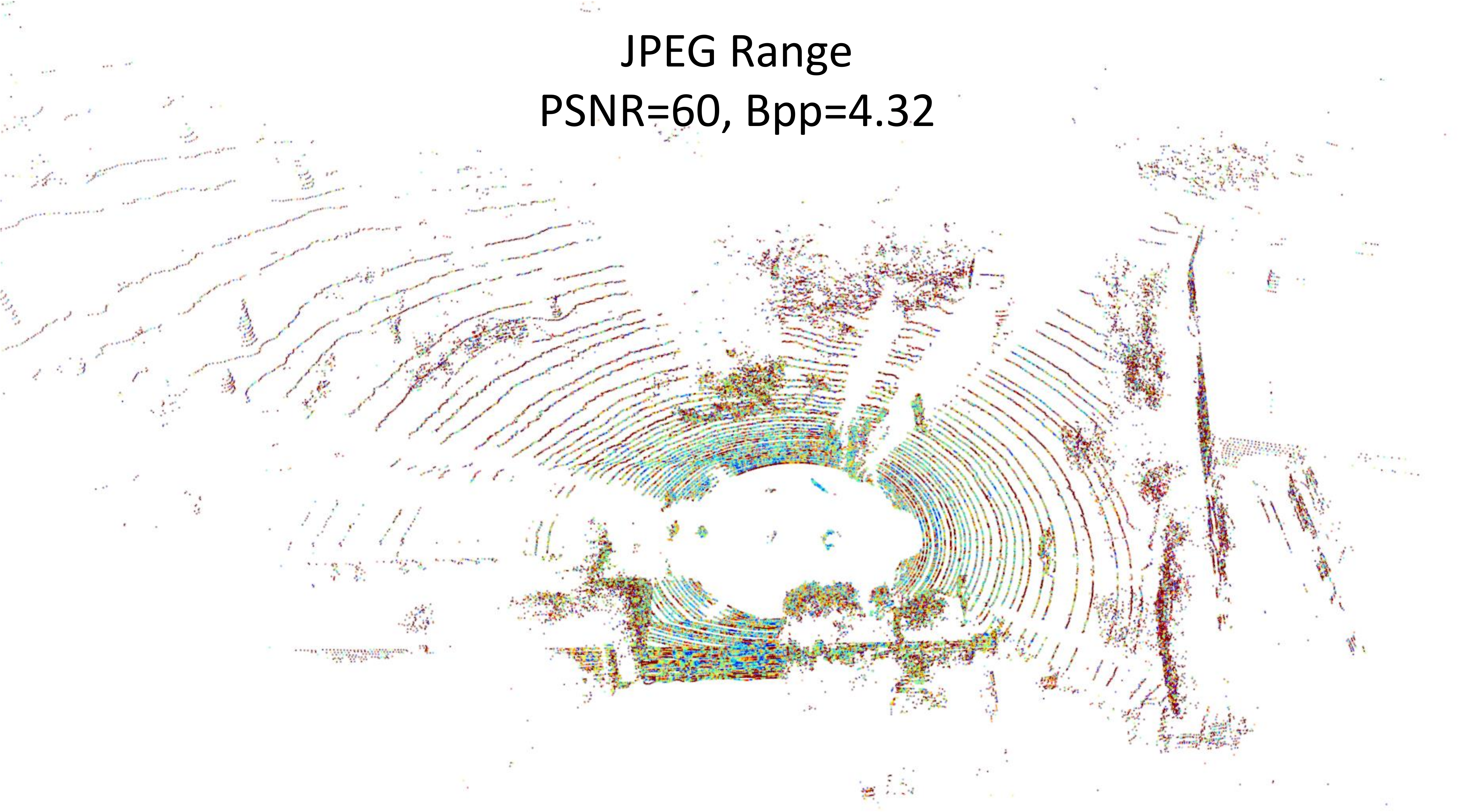}
\end{minipage}
\hfill
\centering
\begin{minipage}[t]{0.19\linewidth}
\centering
\includegraphics[width=1.0\linewidth]{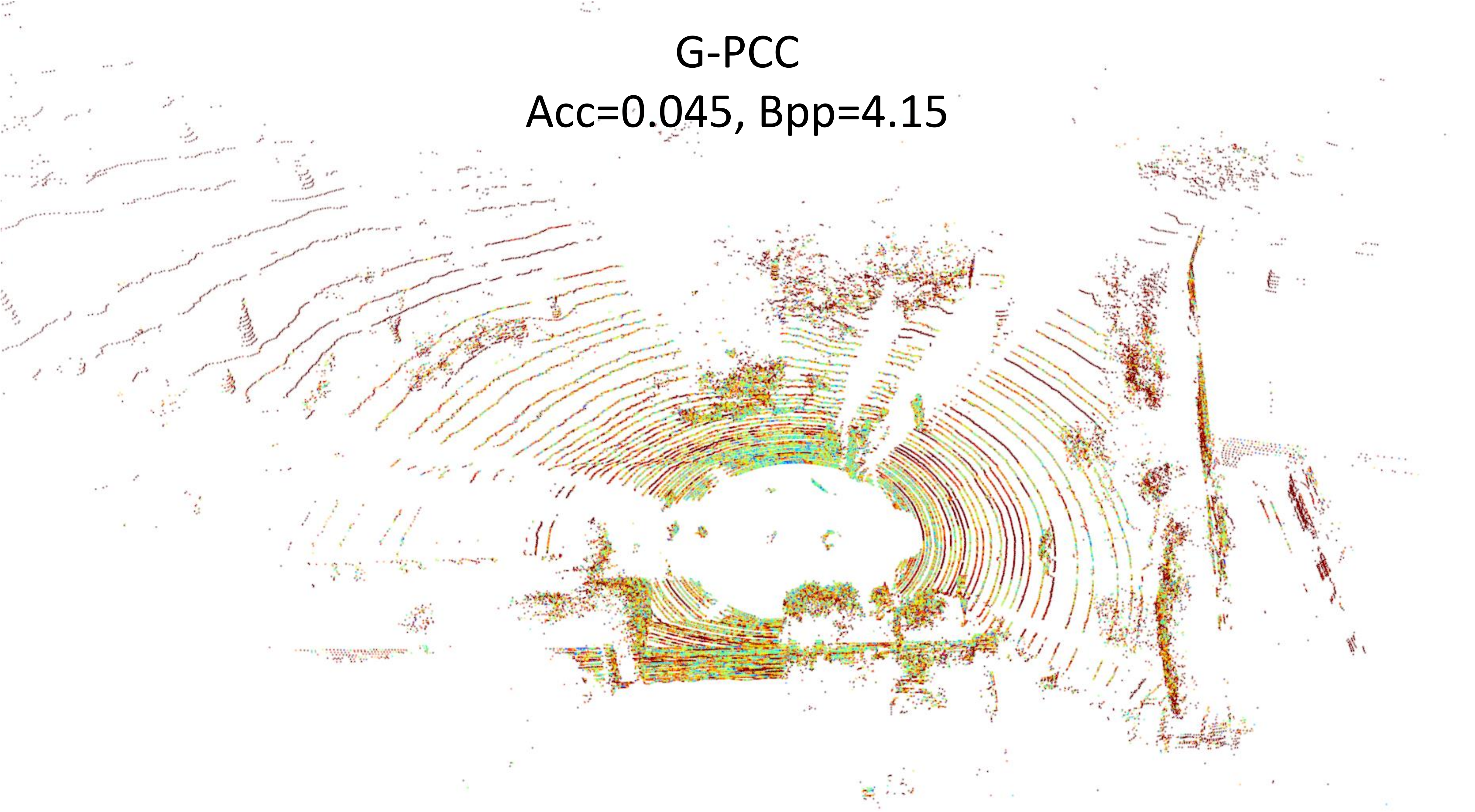}
\end{minipage}
\hspace{0.4em}
\vfill
\begin{minipage}[t]{0.5\linewidth}
\centering
\includegraphics[width=1.0\textwidth]{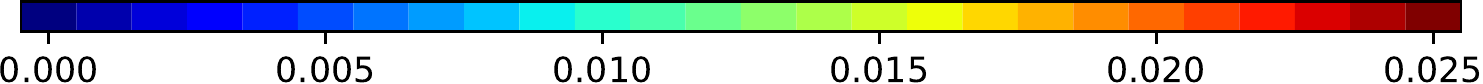}
\end{minipage}
\caption{The qualitative results of our uniform or non-uniform reconstructed point cloud compared with other baseline methods on the KITTI dataset. The color in the point cloud and colorbar is based on the mean symmetric chamfer distance (Eq. \eqref{eq:cd}) between the reconstructed and original point cloud.  }
\label{fig:qualitative}
\vspace{-0.4cm}
\end{figure*}

\textbf{Reconstruction Quality.}
Fig. \ref{fig:compare_1} shows the quantitative results of our method with baseline methods on the KITTI city dataset. It shows that the compression effectiveness and reconstruction quality of our framework is much better than that of the other three baseline methods, because we can control the maximum error of the reconstructed point cloud and the high-level basic compressor can compress the range image better. The reconstruction quality of non-uniform compression is a little less than the uniform compression because the unimportant regions have large error.

\textbf{Downstream Tasks.}
The first row of Fig. \ref{fig:compare_2} shows the evaluation results of the 3D object detection  with  a reconstructed point cloud and raw data. Compared with other methods, our proposed compression framework can reach "lossless" compression and decompression at a lower compressed size (BPP).   
And though the non-uniform framework is not as good as the uniform framework in terms of reconstruction quality, its performance in downstream tasks is better than that of the uniform framework, because the important objects and features are kept after compression and decompression. The performance improvement of the non-uniform framework is more obvious in small object detection like pedestrians or cyclists. The point-lossless feature of our framwork is another advantage in hard object detection.

The second row of Fig. \ref{fig:compare_2} presents the comparative results of A-LOAM on the KITTI sequence $00$ along with different BPPs. The results show that the SLAM performance is less related to the reconstruction quality, which means we can obtain lossless SLAM results with a higher compression ratio and a less-compressed bitrate.

\subsection{Qualitative Results}

In this section, we show the reconstructed point cloud with the original point cloud in Fig. \ref{fig:qualitative}. The color bar shows the error between the reconstructed and original point cloud. The first parameter in the caption is the setting of each method, and the BPP is the bitrate. We can find that when BPP is near 2, the points in reconstructed point cloud of the pedestrians are almost the same as in the original point cloud.

\section{Conclusion}
Our proposed uniform and non-uniform range image-based compression method can be seen as a baseline for large-scale lossless point cloud compression. In downstream tasks such as 3D object detection and SLAM, our method can obtain a compressed bitstream of smaller size than the other baseline methods when the point cloud is lossless. Also, our non-uniform point cloud compression framework can be merged with machine learning for better salience map creation. By assigning a larger accuracy loss to the unimportant areas, we could obtain a higher compression ratio in future work.

\clearpage

\bibliographystyle{IEEEtran}
\bibliography{egbib}
\end{document}